\documentclass[journal,twocolumn]{IEEEtran}
\ifCLASSINFOpdf
\else
\fi
\usepackage{times}
\usepackage{epsfig}
\usepackage{graphicx}
\usepackage{amsmath}
\usepackage{amssymb}
\usepackage{subfigure}
\usepackage{amsmath}
\usepackage{amsthm}
\usepackage{bbm}
\usepackage{url}
\usepackage{cite}
\theoremstyle{definition}
\newtheorem{definition}{Definition}

\newtheorem{theorem}{Theorem}
\DeclareMathOperator*{\argmin}{arg\,min}

\usepackage{multirow}
\usepackage{amsfonts}

\begin{document}
%
\title{Learning Smooth Representation for Unsupervised Domain Adaptation}
%
%
%

\author{Guanyu Cai,
			Lianghua He, 
			Mengchu Zhou,~\IEEEmembership{Fellow,~IEEE},
			Hesham Alhumade,
			and Die Hu
\thanks{This work was supported in part by Joint Funds of the National Science Foundation of China under Grant U18092006, in part by the Shanghai Municipal Science and Technology Committee of Shanghai Outstanding Academic Leaders Plan under Grant 19XD1434000, in part by the Projects of International Cooperation of Shanghai Municipal Science and Technology Committee under Grant 19490712800, in part by the National Natural Science Foundation of China under Grant 61772369, Grant 61773166, Grant 61771144, in part by National Key R\&D Program of China under Grant 2020YFA0711400, in part by Shanghai Municipal Science and Technology Major Project (2021SHZDZX0100), in part Shanghai Municipal Commission of Science and Technology Project(19511132101), in part by the Changjiang Scholars Program of China, in part by the Fundamental Research Funds for the Central Universities.}
\thanks{G. Cai and L. He are with the Department of Computer Science and Technology, Tongji University, Shanghai 201804, China (email: caiguanyu@tongji.edu.cn; helianghua@tongji.edu.cn).}
\thanks{M. Zhou is with the Department of Electrical and Computer Engineering, New Jersey Institute of Technology, Newark, NJ 07102 USA (e-mail: zhou@njit.edu).}
\thanks{H. Alhumade is with the Department of Electrical and Computer Engineering, King Abdulaziz University, Jeddah 21481, Saudi Arabia (email: halhumade@kau.edu.sa)}
\thanks{D. Hu is with the Key Laboratory of EMW Information, Fudan University,
Shanghai 200433, China (e-mail: hudie@fudan.edu.cn).}
}

\maketitle

\begin{abstract}

Typical adversarial-training-based unsupervised domain adaptation methods are vulnerable when the source and target datasets are highly-complex or exhibit a large discrepancy between their data distributions. Recently, several Lipschitz-constraint-based methods have been explored. The satisfaction of Lipschitz continuity guarantees a remarkable performance on a target domain. However, they lack a mathematical analysis of why a Lipschitz constraint is beneficial to unsupervised domain adaptation and usually perform poorly on large-scale datasets. In this paper, we take the principle of utilizing a Lipschitz constraint further by discussing how it affects the error bound of unsupervised domain adaptation. A connection between them is built and an illustration of how Lipschitzness reduces the error bound is presented.  A \textbf{local smooth discrepancy} is defined to measure Lipschitzness of a target distribution in a pointwise way. When constructing a deep end-to-end model, to ensure the effectiveness and stability of unsupervised domain adaptation, three critical factors are considered in our proposed optimization strategy, i.e., the sample amount of a target domain, dimension and batchsize of samples. Experimental results demonstrate that our model performs well on several standard benchmarks. Our ablation study shows that the sample amount of a target domain, the dimension and batchsize of samples indeed  greatly impact Lipschitz-constraint-based methods' ability to handle large-scale datasets. Code is available at \url{https://github.com/CuthbertCai/SRDA}.
\end{abstract}


%
\IEEEpeerreviewmaketitle

\section{Introduction}
%
%
%
%
\IEEEPARstart{U}{nsupervised} domain adaptation (UDA) typically tackles the performance drop once there exists a \emph{dataset shift} between training and testing distributions 
~\cite{krizhevsky2012imagenet,ben2010theory,donahue2014decaf:,pmlr-v119-kumar20c}.  
With the development of computer vision, advanced tasks, such as self-driving cars and robots, need large-scale annotated data. Due to the labor-intensive process of annotation, using synthetic data and computer-generated annotation becomes popular. However, training samples from simulators are synthesized by 3D rendering models whereas testing samples are real-world scenes. The large discrepancy between training and testing distributions and complexity of image information cause the failures of classical UDA models easily 
~\cite{6847217,pan2010domain,7457281,NIPS2006_2983}. 
More powerful and robust UDA algorithms are desiderated to cope with these progressive situations.

Typical UDA algorithms can be divided into two main categories: homogeneous and heterogeneous. The significant difference between them is that the former assumes that the input space of domains is the same while the latter requires no such assumption. For example, the latter can transfer knowledge from a text dataset to an image one whereas homogeneous methods cannot. In this work, we focus on  homogeneous UDA. 

A typical schema of homogeneous UDA was presented in \cite{ben2010theory}.  The divergence between different distributions is first estimated and then an appropriate optimization strategy is introduced to minimize it.
Several UDA methods 
~\cite{pmlr-v37-long15,zellinger2017central,pmlr-v119-liu20m,long2017deep,long2016unsupervised}
are based on Maximum Mean Discrepancy (MMD). Deep kernels were used to estimate MMD between different distributions \cite{pmlr-v37-long15,zellinger2017central} and optimization strategies were proposed to minimize MMD [11]-[13].  Similar to MMD,  a correlation matrix was introduced in~\cite{sun2016deep} to measure the discrepancy between different domains. Using the mutual information as a measurement was introduced in~\cite{pmlr-v119-liang20a}.  
Many other UDA methods utilized a Proxy $\mathcal{A}$-distance~\cite{ben2010theory} to measure  the divergence between source and target distributions. A strategy to control the Proxy $\mathcal{A}$-distance is to find a feature space of examples where both source and target domains are as indistinguishable as possible~\cite{ben2010theory}. To get indistinguishable feature space, an adversarial training strategy~\cite{ganin2016domain} was proposed by Ganin et al. where an auxiliary network tries to distinguish source and target domains while a main network tries to make domains indistinguishable and classify images. Studies~\cite{liu2016coupled,bousmalis2017unsupervised} followed an adversarial training strategy and conducted it on a pixel-level to enhance models' performance. However, this schema faces two issues. First, although various estimation~\cite{ben2010theory,pmlr-v37-long15,zellinger2017central,sun2016deep} 
is proposed to measure the divergence between source and target distributions, their estimation error becomes larger for more complex distributions in general. Second, a large discrepancy exists when both source and target distributions are complex. It is difficult to design an optimization strategy for reducing such a discrepancy. Neither direct minimization
~\cite{pmlr-v37-long15,zellinger2017central,sun2016deep} 
nor adversarial training~\cite{8833506,arjovsky2017towards} have shown stable performance in the cases of complex distributions.


Another schema~\cite{shu2018a} was introduced by virtual adversarial training~\cite{miyato2018virtual} as a regularization method to avoid a gradient vanishing problem of domain adversarial training~\cite{8833506}. Experimental results in 
~\cite{shu2018a,miyato2018virtual,mao2019virtual}
have proved that a local-Lipschitz constraint is effective in UDA and semi-supervised learning. However, they owed its success to a cluster assumption~\cite{grandvalet2004semi-supervised}, i.e.,  a local-Lipschitz constraint helps input samples get divided into clusters. According to the assumption, input samples in the same cluster come from the same category. They did not analyze mathematically how a local-Lipschitz constraint affects the error bound of  a UDA problem. They cannot be applied to large-scale datasets because they ignored the dimension of samples that affects  the error bound of  a UDA problem.  Another problem is that domain adversarial training is still used by them. It is thus possible to cause a gradient vanishing problem~\cite{8833506}, thereby degrading their performance.  The poor performance on large-scale datasets and a gradient vanishing problem prevent previous Lipschitz-constraint-based UDA methods from being applied to real-world scenarios.

Heterogeneous UDA algorithms face a challenge if the source domain and target domain have different features and distributions, especially in cross-modal applications. The generalization error bound for heterogeneous UDA was analyzed in~\cite{JMLR:v20:13-580,9023553}. Several methods were proposed to align feature space and distributions jointly, such progressive alignment~\cite{8475006} and nonlinear matrix factorization~\cite{8725935}. Further, several methods extended  heterogeneous UDA to more settings, where an optimal transport theory was introduced in~\cite{ijcai2018-412} to tackle semi-supervised heterogeneous UDA and fuzzy-relation nets were proposed in~\cite{9172137} to tackle multi-source heterogeneous UDA.

In this paper, we focus on homogeneous UDA and intend to answer why a local-Lipschitz constraint is effective in solving UDA problems and analyze several essential factors that prevent previous Lipschitz-constraint-based methods~\cite{shu2018a,mao2019virtual} from working well on large-scale datasets. According to~\cite{Ben-David2014}, the error bound of a UDA problem is determined by probabilistic Lipschitzness and a constant term. A local-Lipschitz constraint is a special case of probabilistic Lipschitzness such that it helps us tackle UDA problems. However, the methods in~\cite{shu2018a,mao2019virtual} ignored the constant term that would be extremely large when dealing with large-scale datasets. Does such a large term lead to the poor performance of previous Lipschitz-constraint-based methods? This work answers it.

To expand the application scope of Lipschitz-constraint-based methods, the probabilistic Lipschitzness is further extended by proposing a more concise way to achieve Lipschitz continuity in a target distribution through a newly defined concept called \emph{local smooth discrepancy}. An optimization strategy that takes constant factors analyzed in~\cite{Ben-David2014} into consideration is established. It enables our model to cope with large-scale datasets efficiently and stably. Our model contains a feature generator and classifier. The latter tries to classify source samples correctly and detect sensitive target samples that break down a Lipschitz constraint. The former is trained to strengthen the Lipschitzness of these sensitive samples. The defined local smooth discrepancy measures the  Lipschitzness of a target distribution in a pointwise way. Then two specific methods are introduced to compute it. Utilizing it, a detailed optimization strategy is proposed to tackle a UDA problem by considering the effects of the dimension and batchsize of samples and the sample amount of a target domain. This work aims to make the following contributions to advance the field:
\begin{itemize}
\item[1)] A mathematical analysis is for the first time conducted to explain why a local-Lipschitz constraint reduces the error bound of a UDA problem;
\item[2)] A novel and concise approach to achieve probabilistic Lipschitzness is proposed. The concept of a local smooth discrepancy is defined and used to measure Lipschitzness;
\item[3)] An elaborated optimization strategy that takes the dimension and batchsize of samples  and the sample amount of a target domain into consideration is presented. It enables Lipschitz-constraint-based methods to perform effectively and stably on large-scale datasets.
\end{itemize}
In addition, the proposed approach is extensively evaluated on several standard benchmarks. The results demonstrate that our method performs well on all of them. The ablation study is conducted to validate the role of batchsize and dimension of samples and sample amount of a target domain in its performance changes. 

\section{Related Work}
\label{relatedworkd}

\subsection{Distribution-Matching-Based UDA}
Theoretical work~\cite{ben2010theory} confirmed that a discrepancy between source and target distributions causes an invalid model in a target domain. Because the distribution of a domain is difficult to illustrate, an intuitive thought is to match the statistic characteristics of two distributions instead. Maximum mean discrepancy (MMD), which measures expectation difference in reproducing kernel Hilbert space of source and target distributions, has been widely used 
~\cite{pmlr-v37-long15,zellinger2017central,pmlr-v119-liu20m,long2017deep,pan2010domain,long2016unsupervised,7999259}.
Besides MMD-based methods, many other methods utilized different tricks to match distributions. For example,  the covariance of features was utilized in~\cite{sun2016deep} to match different domains. The label information was used in~\cite{9126270} to enhance the distribution matching between different domains in a shared subspace. Using the mutual information as a measurement was introduced in~\cite{pmlr-v119-liang20a}.

\subsection{Adversarial-Training-Based UDA}
Domain-Adversarial Training Network (DANN) was first introduced by Ganin et al.~\cite{ganin2016domain}. An adversarial training strategy to tackle a UDA problem was used in DANN. A domain classifier was introduced to predict which domain a sample is drawn from. Their feature generator was trained to fool its domain classifier such that features from different domains are well-matched. Adversarial Discriminative Domain Adaptation (ADDA)~\cite{tzeng2017adversarial} was proposed by Tzeng et al. that followed this strategy~\cite{ganin2016domain} and introduced an asymmetric network architecture to  obtain a more discriminative representation. Instead of  conducting adversarial training in feature space as DANN and ADDA did, several methods~\cite{liu2016coupled,bousmalis2017unsupervised,murez2018image} implemented adversarial training at pixel level. They tried to generate target images from labeled source images. In this way, a classification model can be trained with labeled target images. Specifically, Pixel-level UDA proposed by Bousmalis et al.~\cite{bousmalis2017unsupervised} followed a training strategy of generative adversarial networks (GANs)~\cite{8039016} and obtained excellent performance on digits datasets. The methods proposed by Murez et al.~\cite{murez2018image} and Liu et al.~\cite{liu2016coupled} introduced a training strategy similar to cycle GANs to improve their models' performance. Besides taking a marginal distribution into consideration, a conditional domain adversarial network (CDAN) proposed by Long et al.~\cite{long2018conditional} adopted the joint distribution of samples and labels in an adversarial training. The label information was used to enhance adversarial training and achieve remarkable results. Specifically, CDAN exploited discriminative information conveyed in the classifier predictions to assist adversarial training. Based on the domain adaptation theory~\cite{ben2010theory}, CDAN gave a theoretical guarantee on the generalization error bound and achieved the best results on five benchmark datasets.

\subsection{Large-Margin-Based UDA}
Several UDA methods~\cite{pmlr-v70-saito17a,saito2018maximum} assumed that classifiers provide adequate information about different distributions. They enforced robust classifiers with large margins. Asymmetric tri-training~\cite{pmlr-v70-saito17a} added two auxiliary classifiers to assist in generating valid pseudo labels for target samples. They constructed decision boundaries with large margins for a target domain with the help of two auxiliary classifiers. The critical effect of large enough margins was also explored by Lu et al.~\cite{8325317}. Maximum Classifier Discrepancy for Domain Adaptation (MCD)~\cite{saito2018maximum} was proposed by Saito et al. and it well answered why the use of matching marginal distributions causes misclassification and why constructing decision boundaries with large margins reduces the error bound of UDA. It also proposed a Siamese-like network and adversarial training strategy to solve a UDA problem. Deep Max-Margin Gaussian Process Approach (GPDA)~\cite{8953535} adopted MCD's principle and introduced a Gaussian process to enhance GPDA's performance. MCD and GPDA were comparable with each other and superior to previous methods on several datasets.

\subsection{Lipschitz-Constraint-Based UDA} 
Recently, a local-Lipschitz constraint [21]-[23]
was introduced into UDA and semi-supervised learning to avoid an unstable training process faced by some adversarial-training-based UDA methods~\cite{8833506}. Compared with MMD-based methods, they got rid of matching different distributions. When aligning distributions, MMD-based methods need to estimate some statistics of both source and target domains. Two problems are confronted in such alignment. First, none of the statistics can describe a distribution perfectly because they represent only part of the distribution. Second, it is hard to estimate statistics of a high-dimensional distribution precisely. However, Lipschitz-constraint-based methods only need to satisfy a Lipschitz constraint in the target domain such that drawbacks of MMD-based methods are avoided. In detail, the virtual adversarial training ~\cite{miyato2018virtual} firstly introduced a Lipschitz constraint into semi-supervised learning. Decision-boundary Iterative Refinement Training with a Teacher (DIRT) ~\cite{shu2018a} explicitly incorporated the virtual adversarial training~\cite{miyato2018virtual} and added the loss of it to the objective function as an additional term. It indeed improved the performance on several benchmark datasets. However, DIRT introduced a cluster assumption~\cite{grandvalet2004semi-supervised} to explain its effectiveness instead of rigorous proof. Meanwhile, an adversarial training in its optimization procedure could lead to a vanishing gradient problem during training~\cite{8833506}.

\subsection{Heterogeneous UDA}
Besides minimizing distribution discrepancy, heterogeneous UDA algorithms have to project cross-modal features into a common feature space. A knowledge transfer theorem and a principal angle-based metric were given in~\cite{9023553}. A fuzzy-relation net~\cite{9172137} was introduced by Liu et al. to expand heterogeneous UDA  to a multi-source setting. A progressive alignment was introduced in~\cite{8475006} and was used to optimize both feature discrepancy and distribution divergence in a unified objective function. A nonlinear matrix factorization was proposed in~\cite{8725935} where nonlinear correlations between features and data instances could be exploited to learn heterogeneous features for different domains. The optimal transport theory was utilized in~\cite{ijcai2018-412} to tackle the semi-supervised  heterogeneous UDA problem.

\section{Preliminary}
\label{prelinary}
In this section, we give a brief description of a UDA problem and define several properties relevant to realizing  a practical UDA algorithm. Moreover, a basic UDA error bound is derived from these properties~\cite{Ben-David2014}\footnote {Note that the theorems assume binary classification ($y\in {0, 1}$). However, they can be directly extended to multi-class settings.}. 

Let $(\chi, \mu)$ be a domain set where $\chi$ denotes a set of samples and $\mu: \chi^{2}\rightarrow\mathbb{R}^{+}$ is a divergence metric over $\chi$. $\mathbb{R}^{+}$ is the set of all non-negative real numbers. In a UDA setup, we denote $P_{S}$ and $P_{T}$ as a source and target distribution, respectively. The marginal distributions of $P_{S}$ and $P_{T}$ over $\chi$ are denoted by $D_{S}$ and $D_{T}$ and their labeling rules are denoted by $l_s: \chi\rightarrow[0, 1]$ and $l_t: \chi\rightarrow[0, 1]$. The goal is to learn a function $l:\chi\rightarrow\{0, 1\}$ which predicts correct labels for samples in $\chi$ with respect to $P_{T}$ over $(\chi\times\{0, 1\})$. Considering that $l_s$ and $l_t$ are defined with a continuous range while $l$ is defined with a concrete range, a common way to map $l_s$ and $l_t$ to a concrete range is to set a threshold $\beta$. If $l_s(x)>\beta$ ($l_t(x)>\beta$), then $x$ is assigned with $1$ else $0$.   For any hypothesis $h:\chi\rightarrow\{0, 1\}$, we define the \textit{error} with respect to $P_{T}$ by $E_{P_{T}}(h)=\mathbb{P}_{(x, y)\sim P_{T}}(y\neq h(x))$ where $\mathbb{P}$ denotes a probability that the label of $x\sim P_{T}$ is different from the output of $h(x)$. Thus, the Bayes optimal error for $P_{T}$ is $E^*(P_{T}):= \min_{h\in\{0, 1\}^{\chi}}E_{P_{T}}(h)$.

Besides basic notations, in the context of realistic UDA problems, there are some properties expressing several conditions of these distributions that enable a practical UDA algorithm~\cite{Ben-David2014}.
\begin{definition}[Covariate shift~\cite{Sugiyama05generalizationerror}]
Source and target distributions satisfy the \textit{covariate shift} property if they have the same labeling rule, i.e., $l_s(x)=l_t(x)$. The common labeling function is denoted as $l=l_s=l_t$.
\label{def1}
\end{definition}
 
\begin{definition}[Probabilistic Lipschitzness~\cite{Ben-David2014}]
Let $\Phi: \mathbb{R}\rightarrow[0, 1]$. $f:\chi\rightarrow\mathbb{R}$ is $\Phi$-$Lipschitz$ w.r.t. distribution $D$ over $\chi$ if, for all $\lambda>0$:
\begin{align}
\mathbb{P}_{x\sim D}[\exists y: |f(x)-f(y)|>\lambda\mu(x,y)]\leq\Phi(\lambda)\label{eq1}
\end{align}
\label{def2}
\end{definition}
This definition generalizes the standard $\lambda$-$Lipschitz$ property where a function $f:\chi\rightarrow\mathbb{R}$ satisfies $|f(x) - f(y)|\leq\lambda\mu(x, y)$ for all $x, y\in\chi$. In the standard $\lambda$-$Lipschitz$ property, if the labeling function is deterministic, just like the goal function $l(x)\in\{0,1\}$ for all $x\in\chi$, the standard property forces $\mu(x,y)\geq1/\lambda$ if $x$ and $y$ belong to different category and $|f(x) - f(y)|\in\{0,1\}$ for all $x$ and $y$. Thus, the standard property is unfit for a concrete range space. However, the probabilistic Lipschitzness just requires the inequality $|f(x) - f(y)|\leq\lambda\mu(x, y)$ to hold only with some probability. Namely,  it does not require all points to hold the inequality. Unlike the standard property that becomes meaningless in a concrete range space because $|f(x) - f(y)|$ is always $0$ or $1$, the probabilistic Lipschitzness forms a constraint over $\mathbb{P}_{x\sim D}[\exists y: |f(x)-f(y)|>\lambda\mu(x,y)]$ that is defined with a continuous range space. The more $y\in\chi$ that satisfies the inequality, the greater the probability. Such relaxation makes  the probabilistic Lipschitzness applicable to the goal function $l(x)\in\{0,1\}$.

\begin{definition}[Weight ratio~\cite{Ben-David2014}]
Let $\mathcal{B}\subseteq2^{\chi}$ be a collection of subsets of $\chi$ measurable with respect to both $D_S$ and $D_T$. For some $\eta>0$ we define the $\eta$-weight ratio of source and target distributions with respect to $\mathcal{B}$ as
\begin{align}
R_{\mathcal{B},\eta}(D_S,D_T)=\inf_{\substack{b\in\mathcal{B}\\
D_T(b)\geq\eta}}\frac{D_S(b)}{D_T(b)},
\end{align}
Further, the \textit{weight ratio} of source and target distributions with respect to $\mathcal{B}$ is defined as
\begin{align}
R_{\mathcal{B}}(D_S,D_T)=\inf_{\substack{b\in\mathcal{B}\\
D_T(b)\neq0}}\frac{D_S(b)}{D_T(b)},
\end{align}
\label{def3}
\end{definition}
A basic observation about UDA  methods is that they can be infeasible when $D_S$ and $D_T$ are supported on disjoint domain regions. To guard against such scenarios, it is common to assume $R_{\mathcal{B}}(D_S,D_T)>0$.

According to Definition~\ref{def1}-\ref{def3}, an error bound was derived in~\cite{Ben-David2014} for a general UDA learning procedure based on the Nearest Neighbor algorithm. Note that $\chi$ should be a fixed-dimension space and $\mathcal{B}$ should be of finite VC-dimension such that the $\eta$-weight ratio can be estimated from finite samples~\cite{Ben-David2014}. In detail, a domain is assumed to be a unit cube $\chi=[0,1]^d$. $\mathcal{B}$ denotes a set of axes aligned rectangles in $[0,1]^d$ and, given some $\gamma>0$, let $\mathcal{B}_{\gamma}$ be a class of axes aligned rectangles with sidelength $\gamma$.

Given a labeled sample batch $S\subseteq(\chi\times\{0, 1\})$,  $S_\chi$ denotes the set of samples without labels of $S$. For any sample $x\in S_{\chi}$, $l^S(x)$ denotes the label of $x$ in $S$ and $N_S(x)$ denotes the nearest neighbor to $x$ in $S$, $N_S(x)=\argmin_{z\in S_\chi}\mu(x,z)$. Define a hypothesis determined by the Nearest Neighbor algorithm as $\mathbb{H}(x)=l^S(N_S(x))$ for all $x\in\chi$.
\begin{theorem}[Error bound~\cite{Ben-David2014}]\label{the1}
For some domain $\chi=[0,1]^d$, some $R>0$ and $\gamma>0$, let $\mathcal{W}$ be a class of pairs $(P_S,P_T)$ of source and target distributions over $(\chi\times\{0, 1\})$ satisfying the covariate shift assumption, with $R_{\mathcal{B}_{\gamma}}(D_S,D_T)\geq R$, and their common labeling function $l:\chi\rightarrow\{0, 1\}$ satisfying the $\Phi$-probabilistic-Lipschitz property w.r.t a target distribution, for some function $\Phi$. Then, for all $m$, and all $(P_S,P_T)\in\mathcal{W}$,
\begin{align}
\mathop{\mathbb{E}}_{S\sim P_S^m}[E_{P_T}(\mathbb{H})]\leq2E^*(P_T)+\Phi(\gamma)+\frac{4\gamma\sqrt{d}}{Rm^{\frac{1}{d+1}}}
\label{eq4}
\end{align} 
where $m$ denotes the size of $S$ containing points sampled i.i.d. from $\chi$.
\end{theorem}

Theorem~\ref{the1} offers a sufficiently tight error bound for practical UDA, although the first term of Theorem~\ref{the1} is twice the Bayes error rate. According to~\cite{ben2010theory,saito2018maximum}, for a practical UDA problem, $E^*(P_T)$ is a constant that is considered sufficiently low to achieve an accurate adaptation with respect to Definition~\ref{def1}. Otherwise, the UDA problem is impractical to solve. The impact of $E^*(P_T)$ is minimal while the other two terms affect the error bound obviously. Practically, several UDA methods~\cite{shu2018a,mao2019virtual} based on Theorem \ref{the1} show their effectiveness on standard benchmarks. It verifies that Theorem \ref{the1}'s error bound is tight enough for a practical UDA problem. Moreover, considering that the second and third terms affect the error bound heavily, they are further optimized in our method. Two techniques are proposed to ensure their small values, thus making the error bound offered by Theorem \ref{the1} tighter  in the task we focus on, i.e., an image classification UDA problem.

Specifically, inspired by Theorem \ref{the1} that small values of the second and third terms lead to a tight error bound, we propose a loss function and an optimization  strategy. First, to decrease $\Phi$ in a deep end-to-end model, we define a local smooth discrepancy to measure the  probabilistic Lipschitzness with a small number of samples, even with a single sample. It is suitable for deep models, because their parameters are updated with respect to a batch of samples in an iteration. This requires a criterion to measure the probabilistic Lipschitzness with limited samples. Local smooth discrepancy satisfies this requirement. Second, the dimension of samples, i.e., $d$ and the sample count of a target domain, i.e., $m$, affect the error bound. Small $d$ and large $m$ are preferred. An optimization strategy with small $d$ is proposed in this work, and an analysis of how $m$ and batchsize affect the performance of Lipschitz-based UDA methods is to be given. The two improvements are not formally defined because they specialize in a more complex and practical setting, i.e., image classification. However, the need to decrease the bound in Theorem \ref{the1} motivates this study.

\section{Limitation of Previous Lipschitz-Constraint-Based Methods}

Previous Lipschitz-constraint-based methods~\cite{shu2018a,mao2019virtual} were inspired by  virtual adversarial training~\cite{miyato2018virtual}. They explicitly incorporated it and added it to a objective function as a regularization term:
\begin{align}
\begin{gathered}
\mathcal L_v(\theta;{\mathcal D_S}, {\mathcal D_T})=\mathop{\mathbb{E}}_{x\sim D_S}[\max_{||r||\leq\epsilon} D_{K}(h_{\theta}(x)||h_{\theta}(x+r))]+\\
\mathop{\mathbb {E}}_{x\sim D_T}[\max_{||r||\leq\epsilon} D_{K}(h_{\theta}(x)||h_{\theta}(x+r))]
\label{eqv}
\end{gathered}
\end{align}
where $h_{\theta}$ denotes a classifier parameterized by $\theta$, $r$ denotes a vector  with the same shape as input sample $x$ and $D_{K}$ is the Kullback-Leibler divergence. The regularization term was proposed to enforce classification consistency within the norm-ball neighborhood of each sample~\cite{miyato2018virtual}. 

DIRT~\cite{shu2018a} gave some intuitive explanations for the effectiveness of locally-Lipshitz constraint. It regarded (\ref{eqv}) as a constraint to satisfy the cluster assumption~\cite{grandvalet2004semi-supervised}. However, as far as we know, no mathematical analysis has been conducted to connect a local-Lipschitz constraint with the error bound of a UDA problem. In this work, a brief analysis of how a local-Lipschitz constraint affects the error bound of UDA is given. According to Theorem~\ref{the1} and Definition~\ref{def2}, it is clear that probabilistic Lipschitzness is crucial to the error bound of UDA. If $(x+r)$ and $h_{\theta}$ are regarded as $y$ and $f$, respectively, $\mathop{\mathbb {E}}_{x\sim D_T}[\max_{||r||\leq\epsilon} D_{K}(h_{\theta}(x)||h_{\theta}(x+r))]$ is to estimate the expectation of maximum divergence between $f(x)$ and $f(y)$. Its meaning is closely correlated with Definition~\ref{def2} except that Definition~\ref{def2} considers all possible $y$ in $D_T$ while $(x+r)$ only denotes a sample within the neighborhood of $x$. If the expectation of maximum divergence between $f(x)$ and $f(y)$ is small enough, the probability of $[\exists y: |f(x)-f(y)|>\lambda\mu(x,y)]$ is small.  Therefore, once minimizing (\ref{eqv}), $\Phi(\gamma)$ in (\ref{eq4}) is approximately minimized and the error bound of UDA is also reduced.

Although a local-Lipschitz constraint helps reduce the error bound of UDA, there are several issues we need to overcome. The first one is that despite Theorem~\ref{the1} and Definition~\ref{def2} give us the intuition that local-Lipschitz constraint helps reduce the error bound of UDA, they are not derived from a deep learning model whereas recent UDA methods~\cite{shu2018a,mao2019virtual} are mainly based on deep learning models. An explanation is needed to connect a deep end-to-end model with the local-Lipschitz constraint. Secondly, previous Lipschitz-constraint-based methods use an adversarial training strategy. As shown in~\cite{8833506}, adversarial training in UDA methods can easily cause a gradient vanishing problem. It thus hurts the stability of such methods. Thirdly, besides $\Phi(\gamma)$ and $E^*(P_T)$, Theorem~\ref{the1} contains a constant term. Previous methods~\cite{shu2018a,mao2019virtual} all ignored it but it may increase to an extremely large value when a large-scale dataset is processed. This term prevents previous Lipschitz-constraint-based methods from working well on large datasets.

In this work, we analyze and solve the above three problems. A brief explanation is given to answer why Theorem~\ref{the1} holds with a deep learning model. The adversarial training strategy is removed in our proposed method and a concise minimization procedure is utilized to maintain stable performance. Another important contribution of our work is that we answer how the constant term affects Lipschitz-constraint-based methods' performance and extend such methods to work well on large datasets.

\section{Learning Smooth Representation}
\label{ourmethod}

\begin{figure*}[htp]
\centering
\includegraphics[width = \textwidth]{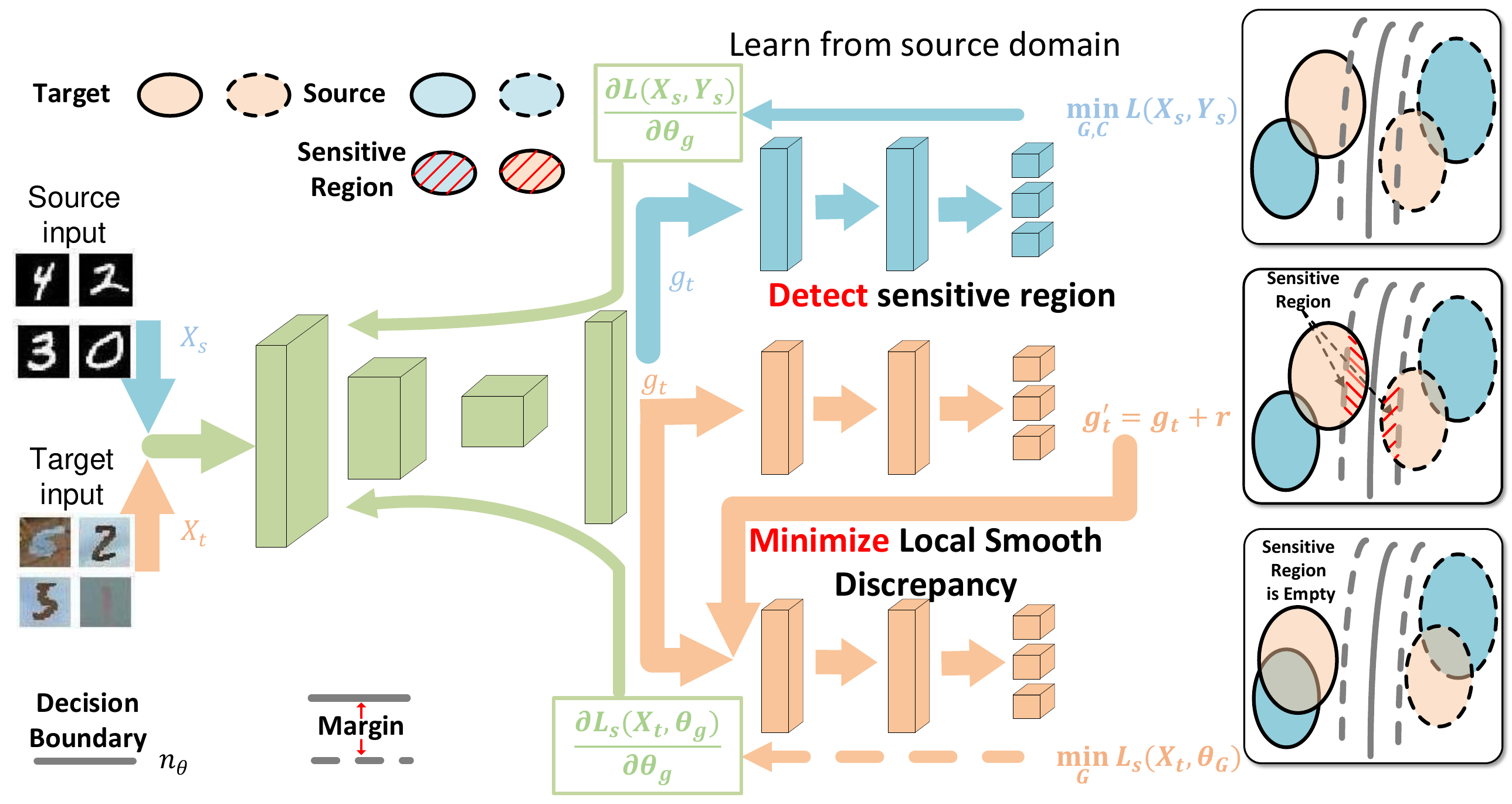}
\caption{A visual illustration of how the proposed method achieves adaptation. The decision boundary is determined by learning from a source domain. Then, the high-density region is detected by seeking sensitive samples of a target distribution. Finally, smooth representation is learned by minimizing local smooth discrepancy. Dashed lines indicate that gradients are not applied.}
\label{fig:motivation}
\end{figure*} 

In Theorem~\ref{the1}, the error bound of a general UDA learning algorithm is bounded by three terms. The first one, $E^*(P_T)$, refers to a Bayes optimal error for $P_T$. It is a value discussed in the theoretical analysis which is impossible to obtain in practical settings. Therefore, the other two terms are the priorities in this paper. The second term, $\Phi(\lambda)$, refers to the degree of how target samples satisfy  probabilistic Lipschitzness. Because $\mathop{\mathbb{E}}_{S\sim P_S^m}[E_{P_T}(\mathbb{H})]$ is positively related to $\Phi(\lambda)$, our goal is to decrease $\Phi(\lambda)$, i.e., strengthen a probabilistic-Lipschitz property with respect to a target distribution. The last term is relevant to multiple factors, for example, the lower bound $R$ for the weight ratio of source and target distributions, the dimension $d$ of domain $\chi$ and the size $m$ of $S$. Note that the weight ratio $R_{\mathcal{B}_{\gamma}}(D_S,D_T)$ is predetermined for a specific pair $(P_S,P_T)$. Therefore, it is impracticable to optimize its lower bound $R$. To achieve a tighter error bound of a target distribution, it is reasonable to explore appropriate $d$ and $m$. In summary, two principles, i.e., strengthening probabilistic-Lipschitz property and searching proper $d$ and $m$, guide us to tackle a UDA problem thoroughly. Similar analyses are deduced from~\cite{Ben-David2014}, while no applicable algorithm is proposed in~\cite{Ben-David2014}. In this paper, we adopt the well-established principles and innovatively extend them to a feasible and effective UDA algorithm.

\subsection{Deep End-to-End Model}
\label{deep}

Although the proposed principles are reasonable, the first problem we should deal with is how to design a deep end-to-end model aligned well with them. Deep neural networks with numerous parameters optimized by gradient-based algorithms can well handle large-scale datasets~\cite{krizhevsky2012imagenet}. An error bound with respect to a deep neural network is  derived as follows.

Given a deep neural network $n_\theta$ parameterized by $\theta$ whose input samples are $x\in\chi$, if the output range space of $n_\theta$ is defined in $[0, 1]$, $n_\theta$ satisfies the definition of labeling rules just like $l_s$ and $l_t$. For any sample $x\in S_{\chi}$, $n_\theta(x)$ is used to replace $l^S(x)$ to denote the label of $x$ in $S$ and $N_S(x)$ denotes the nearest neighbor to $x$ in $S$, i.e., $N_S(x)=\argmin_{z\in S_\chi}\mu(x,z)$. We make a hypothesis determined by the Nearest Neighbor algorithm as $\mathbb{H^\prime}(x)=n_\theta(N_S(x))$ for all $x\in\chi$.

 Because $n_\theta$ meets the definition of labeling rules and Theorem~\ref{the1} is derived based on the labeling rules, an error bound based on $n_\theta$ is given as:
\begin{align}
\mathop{\mathbb{E}}_{S\sim P_S^m}[E_{P_T}(\mathbb{H^\prime})]\leq&2E^*(P_T)+\Phi(\gamma)+\frac{4\gamma\sqrt{d}}{Rm^{\frac{1}{d+1}}}\label{eqdeep}
\end{align}
Thus, the local-Lipschitz constraint can be applied to a deep end-to-end model.

\subsection{Local Smooth Discrepancy}

Besides extending Theorem~\ref{the1} to a deep end-to-end model, how to decrease $\Phi(\lambda)$ is another crucial problem. According to (\ref{eqdeep}), the basic UDA error bound is positively related to $\Phi(\lambda)$.  Because $\Phi(\lambda)=\sup_{x\in D}\mathbb{P}_{x\sim D}[\exists y: |f(x)-f(y)|>\lambda\mu(x,y)]$, minimizing $\mathbb{P}_{x\sim D}[\exists y: |f(x)-f(y)|>\lambda\mu(x,y)]$  for all $x\in D$ decreases $\Phi$, thus, decreases the error bound.

\begin{figure}[h]
  \centering
  \subfigure[Isotropic plan]{
    \label{fig:isoplan} 
    \includegraphics[width=0.48\columnwidth]{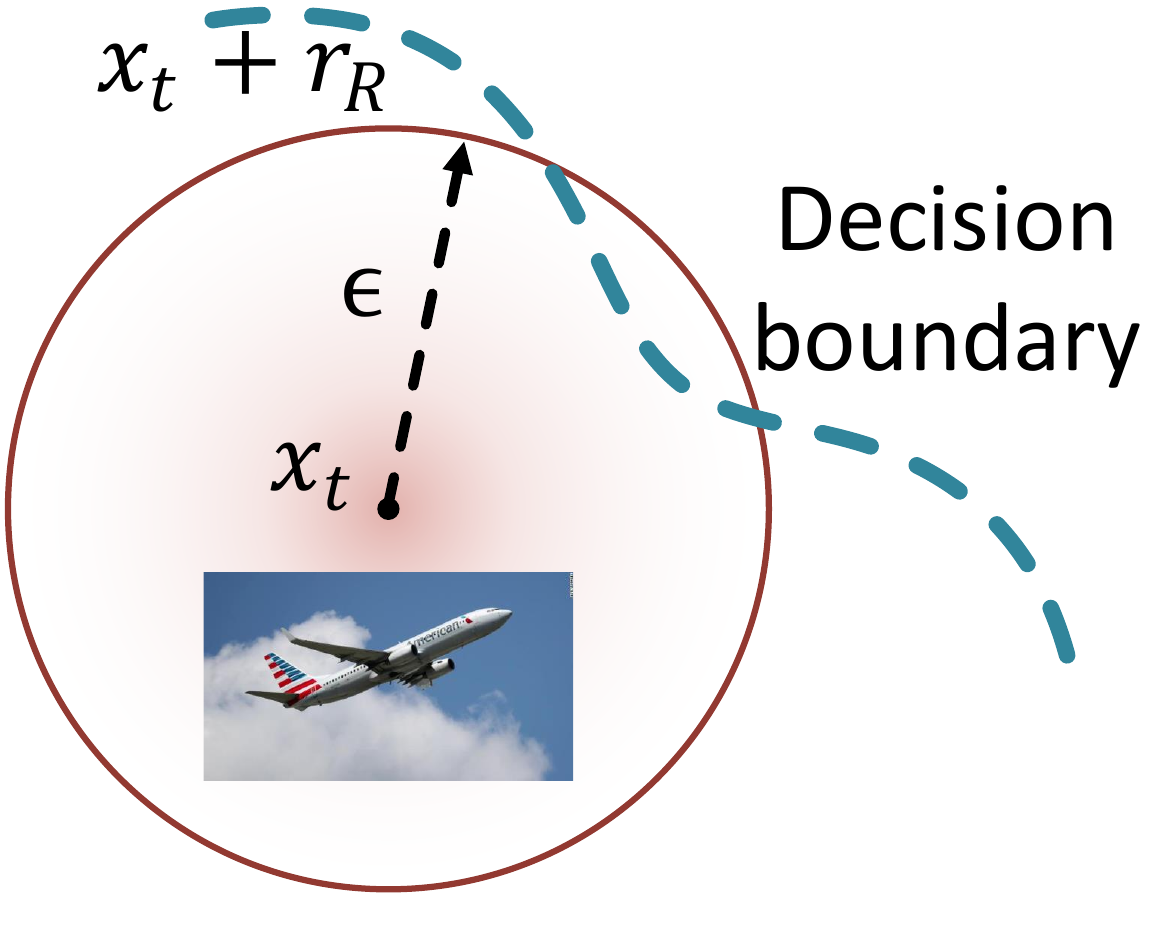}}
  \subfigure[Anisotropic plan]{
    \label{fig:anisoplan} 
    \includegraphics[width=0.48\columnwidth]{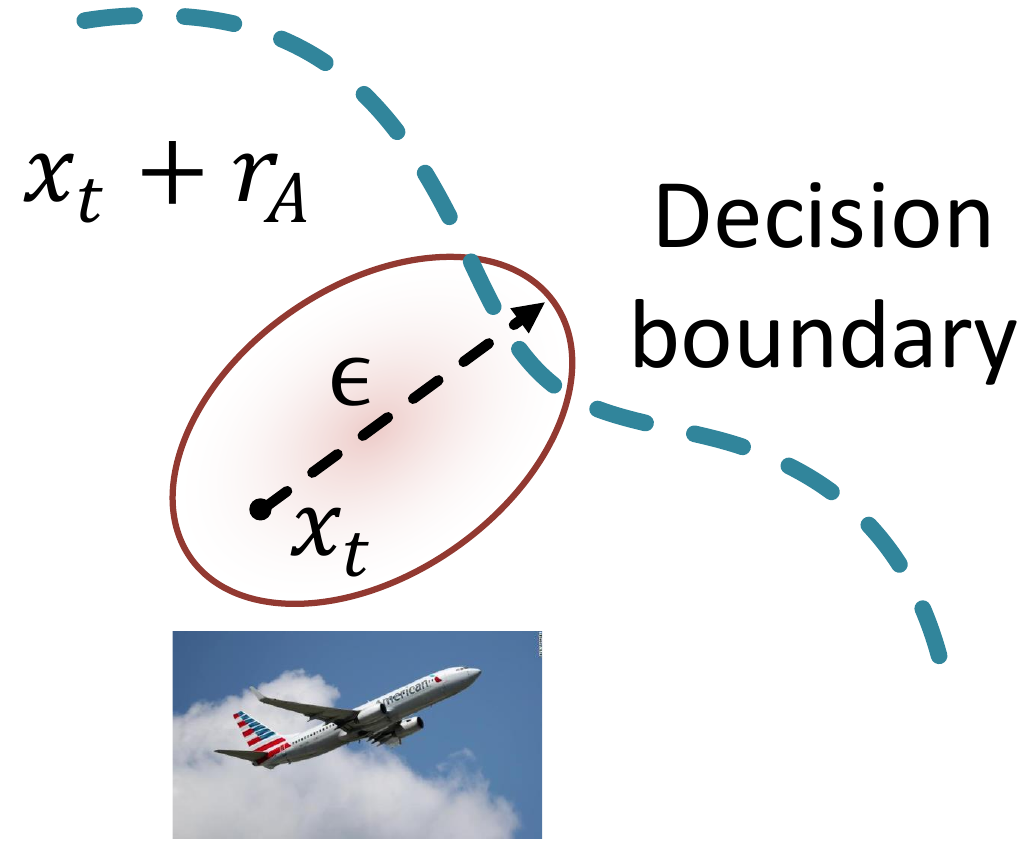}}
\caption{A brief demonstration of isotropic and anisotropic methods.}
  \label{fig:plans} 
\end{figure} 

A naive estimation for $\mathbb{P}_{x\sim D}[\exists y: |f(x)-f(y)|>\lambda\mu(x,y)]$ is $\mathbb{E}_{x\sim D}[\frac{\mathbbm{1}[\exists y:|f(x)-f(y)|>\lambda u(x,y),y\sim D]}{\mathbbm{1}[\exists y:|y\sim D]}]$. $\mathbbm{1}[statement]$ is an indicator function that returns 1 if $statement$ is true, and 0 otherwise. This estimation needs to sample many $y\in D$ to make sure the estimation with low variance. In a deep end-to-end model, if a bunch of $y$ is sampled to calculate $|f(x)-f(y)|$, it requires multiple forward propagations for a batch. According to~\cite{grandvalet2004semi-supervised}, a replacement called local Lipschitz property for Lipschitz continuity is proposed. It assumes that every $x\in D$ has a neighborhood $U$ such that $f$ is Lipschitz continuous with respect to $x$ and points in $U$. According to the local Lipschitz property, we propose a concept called a {\bf local smooth discrepancy} (LSD) to measure the degree that a sample $x$ breaks down the local Lipschitz property in a pointwise way:
\begin{align}
\begin{gathered}
L_s(x,\theta)=D(n_\theta(x+r),n_\theta(x)),\;||r||\leq\epsilon \label{eq5}
\end{gathered}
\end{align}
where $r$ is a random vector with the same shape as $x$. $D(\cdot,\cdot)$ is a discrepancy function that measures the divergence between $(x+r)$ and $x$, and $\epsilon$ denotes the maximum norm of $r$. In $L_s(x,\theta)$, a sample $x$ adds $r$ to detect another sample in $x$'s neighborhood. $\epsilon$ controls the range of $x$'s neighborhood. As for the choice of $D(\cdot,\cdot)$, we employ a cross-entropy loss function in all experiments.

Although the proposed (\ref{eq5}) looks similar to (\ref{eqv}), there are two essential different points. Firstly, according to Theorem~\ref{the1}, $\Phi(\lambda)$ is estimated with respect to a target distribution. Thus, (\ref{eq5}) is only conducted for target samples whereas (\ref{eqv}) applies to both source and target distributions. Our proposed LSD avoids introducing noise from a source distribution and reduces computational cost. Moreover, (\ref{eqv}) requires to seek a point that causes Kullback-Leibler divergence to be largest in a norm-ball neighborhood of each sample. Its intuition is that if the worst point in the neighborhood is enforced to satisfy Lipschitz constraint, the whole neighborhood satisfies Lipschitz constraint. However, our intuition is that points in the neighborhood of the original target sample are sampled to estimate the probability of $n_\theta$ breaking down a Lipschitz constraint. Therefore, (\ref{eq5}) involves every possible point in the neighborhood instead of a single point.

There is another essential point we should pay attention to. Specifically, $r$ is limited only by its norm in (\ref{eq5}) and its direction is ignored. In fact, the goal of adding $r$ includes detecting sensitive samples that do not satisfy a local Lipschitz property. If all $x+r$ belong to the same category as $x$, it means that the direction of $r$ could not detect sensitive samples. In this condition, sensitive samples are not modified to ensure $n_{\theta}$ to be local Lipshitz continuous. To solve this problem, we propose two methods to produce $r$, an isotropic one and an anisotropic one as shown in Fig.~\ref{fig:plans}.

{\bf Isotropic Method:} In this method, $r$ is drawn from a Gaussian distribution and normalized to satisfy $||r||\leq\epsilon$. The formula of LSD is modified into:
\begin{align}
\begin{gathered}
L_s(x,\theta)=D(n_\theta(x+r_{R}),n_\theta(x)),\\
r_{R}=\epsilon\frac{m}{||m||_2},\;m\sim N(0,1)\label{eq6}
\end{gathered}
\end{align}
where $r_{R}$ denotes a noise vector sampled from a Gaussian distribution.

{\bf Anisotropic Method:} This method only looks for $r$ that leads $x+r$ with different labels from $x$ and ignores $r$ in other directions. It can be considered as a biased estimation of the probability that target samples break down a Lipschitz constraint, because the samples that satisfy a Lipschitz constraint are neglected. To implement this method, an insight from an adversarial attack algorithm~\cite{goodfellow2014explaining} is taken. It applies a certain hardly perceptible perturbation, which is found by maximizing the model's prediction error, to an image to cause the model to misclassify~\cite{goodfellow2014explaining}. Similar to its goal, the anisotropic method tries to search for a perturbation to cause a model to output different labels for an image. However, true labels are needed in adversarial attack algorithms. In a UDA problem, true labels for a target domain are not available. Therefore, several modifications are introduced in a traditional adversarial attack method. In detail, it approximates an adversarial perturbation by~\cite{goodfellow2014explaining}:
\begin{align}
r_{A}=\epsilon\frac{m}{||m||_2},\;m = \nabla_{x}D(n_\theta(x),y)
\end{align}
where $r_{A}$ denotes an anisotropic noise that leads to a wrong label. To get rid of the constraint of label information, $r_{A}$ is approximated by:
\begin{align}
\begin{gathered}
r_{A}\approx\epsilon\frac{m}{||m||_2},\\
m = \nabla_{x}D(n_\theta(x),\;onehot(n_\theta(x)))\label{eq8}
\end{gathered}
\end{align}
where $onehot$ denotes a function that transforms the softmax output of $n_\theta$ to a one-hot vector. (\ref{eq8}) computes gradients of $x$ and replaces $y$ with $onehot(n_\theta(x))$. These modifications result in a new LSD with anisotropic noise:
\begin{align}
L_s(x,\theta)=D(n_\theta(x+r_{A}),n_\theta(x))\label{eq9}
\end{align}

Both the isotropic and anisotropic methods are proposed to detect sensitive samples to estimate the probability of original target samples breaking down a Lipschitz constraint. However, they are different as shown in Fig.~\ref{fig:plans}. The former, which is an unbiased estimation method, detects all directions around a target sample. It ensures a robust training procedure. However, the latter attempts to detect more sensitive samples and ignores many points around the original sample. It fits the situation that sensitive samples are hard to seek. Meanwhile, because the softmax output of $n_{\theta}$ is regarded as a pseudo label, it can find a wrong direction and result in a vulnerable training procedure.

\subsection{Optimization Strategy}

By optimizing the proposed LSD,  probabilistic Lipschitzness can be satisfied. However, how to optimize it and what factors should be concerned to constrain the third term in Theorem~\ref{the1} need a detailed study.

Basically, either (\ref{eq6}) or (\ref{eq9}) can be adopted as a loss function and use a gradient-based algorithm for optimization. However, there is an essential factor deserved to be considered. The dimension $d$ of domain affects the error bound sharply. For a UDA problem on different image domains, $d$ potentially varies over a large range of values because the size of images is totally different. For instance, when tackling a typical task that adapts a model trained on the MNIST dataset~\cite{726791} to perform well on USPS dataset~\cite{291440}, each image's size is $28\times28$ while the size of an image in VisDA could be $352\times311$. Despite setting other hyper-parameters in the two tasks to be the same, the constant term in Theorem~\ref{the1} varies a lot. Especially, for a large-scale dataset, a large image size results in a large constant term. Considering that a large $d$ leads to a loose error bound, a noise $r$ is generated in feature space instead of an image level such that $d$ is decreased acutely. (\ref{eq6}) and (\ref{eq9}) are thus revised into:
\begin{align}
\begin{gathered}
L_s(x,\theta)=D(C(G(x)+r_{R}),C(G(x))),\\
r_{R}=\epsilon\frac{m}{||m||_2},\;m\sim N(0,1)\label{eq10}
\end{gathered}
\end{align}
\begin{align}
\begin{gathered}
L_s(x,\theta)=D(C(G(x)+r_{A}),C(G(x)))\\
r_{A}\approx\epsilon\frac{m}{||m||_2},\\
m = \nabla_{g}D(C(g),\;onehot(C(g))),\;g=G(x)\label{eq11}
\end{gathered}
\end{align}
where $G$ and $C$ denote a feature extractor and classifier in $n_\theta$, respectively. Their parameters are denoted as $\theta_G$ and $\theta_C$.

Then an optimization strategy is proposed in feature space in three steps. Firstly, $G$ and $C$ are trained in a source domain.
\begin{align}
\begin{gathered}
\min_{G,C}{\cal L}(X_s,Y_s),\\
{\cal L}(X_s,Y_s)=\mathbb{E}_{x,y\sim P_S}[\sum_{k=1}^{K}\mathbbm{1}[k=y]\cdot log(C(G(x)))]\label{eq12}
\end{gathered}
\end{align} 
where $\mathbbm{1}[\cdot]$ is an indicator function, and $K$ denotes the number of classes in a task. Then, sensitive samples that break down a local Lipschitz property are produced. Note that in Theorem~\ref{the1}, a Lipschitz property constraint is satisfied w.r.t a target distribution such that we focus on target samples in the second step. In our work, a sensitive sample $g^{\prime}_t$ is generated in the output space of $G$:
\begin{align}
g^{\prime}_t= g_t+r\label{eq13}
\end{align}
where $g_t=G(x_t)$, and $r$ is a general notation for the adding noise. In practice, we set $r=r_{R}$ for an isotropic method or $r=r_{A}$ for an anisotropic method. Finally, $G$ is trained to minimize $L_s$ for target samples. Only parameters of $G$ are updated in this step. $G$ is trained to project $g^{\prime}_t$ to the same category with $g_t$:
\begin{align}
\begin{gathered}
\min_{G}L_s(X_t,\theta_G),\\
L_s(X_t,\theta_G)={\mathbb E}_{x\sim P_t}D(C( g^{\prime}_t),C(g_t))\label{eq14}
\end{gathered}
\end{align}
where $\theta_G$ denotes the aparameters of $G$, and $D(\cdot,\cdot)$ denotes a cross-entropy loss function. (\ref{eq12})-(\ref{eq14}) are repeated in the optimization schedule as shown in Fig.~\ref{fig:motivation}.

An intuitive understanding of our optimization strategy is shown in Fig.~\ref{fig:motivation}. In detail, $n_\theta$, which is well-trained in a source domain, classifies source samples correctly. For a robust model, there is a large margin between a decision boundary and samples. It ensures that a local Lipschitz property is well-satisfied. However, a dataset shift may push some target samples to cross the boundary. They are misclassified and the local Lipschitz constraint is broken down. To obtain $n_\theta$ that works well in a target domain, $G$ needs to project $x\sim P_T$ into feature space away from the decision boundary.

In our strategy, a large margin is formed between target samples and a decision boundary obtained in a source domain. The proposed optimization strategy detects samples close to a boundary and forces $G$ to project them far away from the boundary. It makes the representation of target distribution become ``smooth" gradually, implying that a local Lipschitz constraint is ensured. When the optimization procedure ends, a large margin between a target distribution and decision boundary is formed.

Another factor, i.e., the sample amount of a target domain $m$  also affects the constant term in Theorem~\ref{the1}. Briefly, a large $m$ is recommended. According to Theorem~\ref{the1}, it is clear that when $m$ increases, the constant term $\frac{4\gamma\sqrt{d}}{Rm^{\frac{1}{d+1}}}$ decreases. Thus, the fact that larger $m$ causes smaller $\frac{4\gamma\sqrt{d}}{Rm^{\frac{1}{d+1}}}$ suggests us to set a large $m$ during the training period. In practice, if there is only a small number of samples in the target domain, Lipschitz-constraint-based methods cannot ensure excellent performance, to be shown in our experiments.

Besides $d$ and $m$, the batchsize during training is also a critical factor according to our experimental results. Recent studies~\cite{brock2018large, He_2019_CVPR} have shown that a large batchsize improves the performance of a deep neural network. They concern about image classification and GAN models. To our knowledge, this work is the first study to conclude that a large batchsize is beneficial to UDA problems. To verify our conclusion, an ablation study is conducted next.

\section{Experiments and Discussion}
\label{experiments}

To verify the effectiveness of the proposed method, i.e., learning {\bf S}mooth {\bf R}epresentation for unsupervised {\bf D}omain {\bf A}daptation (SRDA), several classification experiments are conducted on standard datasets. First, the datasets of experiments are introduced. Second,  SRDA are tested on all datasets to show its outstanding performance in comparison with other UDA methods. Then, an ablation study is conducted to analyze how the sample amount of a target domain, and the dimension and batchsize of samples affect Lipschitz-based methods' performance and discuss the effectiveness and robustness of SRDA.

\subsection{Datasets}

{\bf Digits:} Digits datasets include MNIST~\cite{726791}, USPS~\cite{291440}, Synthetic Traffic Signs (SYNSIG)~\cite{10.1007/978-3-319-02895-8_52}, Street View House Numbers (SVHN)~\cite{37648} and German Traffic Signs Recognition Benchmark (GTSRB)~\cite{6033395}. Specifically, MNIST, USPS and SVHN consist of 10 classes, whereas SYNSIG and GTSRB consist of 43 classes. We set four transfer tasks: SVHN$\rightarrow$MNIST, SYNSIG$\rightarrow$GTSRB, MNIST$\rightarrow$USPS and USPS$\rightarrow$MNIST.

{\bf VisDA:} VisDA~\cite{peng2017visda:} is a more complex object classification dataset. It contains more than 280K images belonging to 12 categories. These images are divided into training, validation and test sets. There are 152,397 training images synthesized by rendering 3D models. The validation images are collected from MSCOCO~\cite{lin2014microsoft} and amount to 55,388 in total. This dataset requires an adaptation from synthetic-object to real-object images. We regard the training set as a source domain and the validation set as a target domain. 

{\bf Office-31:} Office-31~\cite{10.1007/978-3-642-15561-1_16} is a small-scale dataset that comprises only 4,110 images and 31 categories collected from three domains: AMAZON (A) with 2,817 images, DSLR (D) with 498 images and WEBCAM (W) with 795 images. We focus on the most difficult four tasks: A$\rightarrow$D, A$\rightarrow$W, D$\rightarrow$A and W$\rightarrow$A. 
\begin{table*}[htbp]
\centering
\caption{Classification accuracy percentage of digits classification experiment among all four tasks. The first row corresponds to the performance if no adaption is implemented. We evaluate three SRDA models with different methods for adding noise. SRDA* denotes the models that are optimized in an image level. The results are cited from each study.}
\setlength{\tabcolsep}{10mm}{
\begin{tabular}{l|cccc}
\hline\hline
\multirow{3}{*}{Method}& SVHN &SYNSIG & MNIST & USPS\\
&$\rightarrow$ & $\rightarrow$ & $\rightarrow$ & $\rightarrow$\\
&MNIST & GTSRB &USPS &MNIST\\
\hline
Source Only& 67.1 & 85.1 & 76.7 & 63.4\\
\hline
DAN~\cite{pmlr-v37-long15}& 71.1 & 91.1 & - & - \\
DANN~\cite{ganin2016domain}& 71.1 & 88.7 & 77.1 & 73.0 \\
DSN~\cite{NIPS2016_6254}& 82.7 & 93.1 & 91.3 & - \\
ADDA~\cite{tzeng2017adversarial}& 76.0 & - & 89.4 & 90.1\\
CoGAN~\cite{liu2016coupled}& - & - & 91.2 & 89.1\\
ATDA~\cite{pmlr-v70-saito17a}& 86.2 &  96.2 &- & - \\
ASSC~\cite{Haeusser_2017_ICCV}& 95.7 & 82.8 & - & -\\
DRCN~\cite{DBLP:conf/eccv/GhifaryKZBL16}& 82.0 & - & 91.8 & 73.7\\
MCD~\cite{saito2018maximum}&96.2 & 94.4 & 94.2 & 94.1\\
\hline
Source Only&82.4&88.6&-&-\\
\hline
VMT~\cite{mao2019virtual}&{\bf 99.4} & - & - & -\\
VADA~\cite{shu2018a}&94.5 & 99.2 & - & -\\
DIRT~\cite{shu2018a}&{\bf 99.4} & {\bf 99.6} & - & -\\
\hline
Source Only& 67.1 & 85.1 & 76.7 & 63.4\\
\hline
SRDA$^F$&95.96 & 90.87 & 85.00 & 95.78\\
SRDA$^{F*}$&22.70&not converge&32.73&85.37\\
SRDA$^V$& 98.90& 92.44& 84.64&95.49 \\
SRDA$^{V*}$&89.47&29.04&88.49&92.17\\
SRDA$^G$&  99.17 & 93.61 & {\bf 94.76}& 95.03\\
SRDA$^{G*}$&89.51&49.86&93.25&{\bf 95.95}\\
\hline\hline
\end{tabular}}
\label{tab:digits}
\end{table*}

Please note that Office-31 is so small-scale dataset that small $m$ leads to a loose error bound for Office-31. By comparing results on Office-31 with those on VisDA, an analysis of how $m$ affects the performance of Lipschitz-constraint-based methods is conducted.

\subsection{Implementation Detail}

The first practical issue is that how to decide which layer to conduct optimization, because LSD is minimized in feature space in the proposed optimization strategy. In all experiments, a general principle is adopted that the last convolutional layer should be chosen to conduct optimization no matter what the network backbone is. For example, if the network backbone is ResNet50~\cite{7780459}, $G$ includes all convolutional layers (49 layers) and $C$ includes a dense layer. Therefore, the number of parameters of $G$ and $C$ varies as the network backbone changes. This principle is reasonable because a small $d$ leads to a low error bound for Lipschitz-constraint-based methods. In the last convolutional layer, $d$ reaches the minimum. The reason for not considering dense layers is that features after dense layers form a cluster if they belong to the same category. A large margin is formed between two different clusters. It is hard to detect sensitive samples because a feature vector with an added noise is classified as the original category.  

{\bf Digits:} For a fair comparison, we follow the network backbone in MCD~\cite{saito2018maximum} and ADDA~\cite{tzeng2017adversarial}. Hyper-parameter $\epsilon$ is set to 0.5 and the learning rate is set to $1e^{-3}$. Batchsize is set to 128 in all tasks and all models are trained for 150 epochs.

Both isotropic and anisotropic methods are implemented. For the former, noise is sampled from a standard Gaussian distribution. For the latter, sensitive samples are generated in two different ways.  Two classical adversarial attack algorithms are chosen, namely Fast Gradient Sign Method (FGSM)~\cite{goodfellow2014explaining} and Virtual Adversarial Training (VAT)~\cite{miyato2018virtual}, to produce noise. FGSM~\cite{goodfellow2014explaining} needs true labels to execute a backpropagation to compute gradients. Instead, pseudo labels are used to replace them. 

{\bf VisDA:} Both isotropic and anisotropic methods are implemented. All models are trained for 15 epochs and batchsize is  32. Learning rate is $10^{-4}$ and hyper-parameter $\epsilon$ is set to 0.5. The backbone network is ResNet101~\cite{7780459} that is the same with MCD~\cite{saito2018maximum} and GPDA~\cite{8953535}.

{\bf Office-31:}  Both the isotropic and anisotropic methods are implemented. For the former, noise is sampled from a standard Gaussian distribution. For the latter, FGSM~\cite{goodfellow2014explaining} is used to produce noise. All models are trained for 50 epochs and batchsize is set to 32. Learning rate is set to $10^{-3}$ and hyper-parameter $\epsilon$ is set to 0.5 for all four tasks.  The backbone network is ResNet101~\cite{7780459}.

\subsection{Model Evaluation}

\subsubsection{Digits Classification}
Results of the digits classification experiment are shown in Table~\ref{tab:digits}. We call SRDA models that generate noise from FGSM, VAT and a random Gaussian distribution as SRDA$^F$, SRDA$^V$ and SRDA$^G$, respectively.  We compare them with other UDA algorithms such as DAN~\cite{pmlr-v37-long15}, DANN~\cite{ganin2016domain}, Domain Separation Network (DSN)~\cite{NIPS2016_6254}, ADDA~\cite{tzeng2017adversarial}, Coupled Generative Adversarial Network (CoGAN)~\cite{liu2016coupled}, Asymmetric Tri-training for Unsupervised Domain Adaptation (ATDA)~\cite{pmlr-v70-saito17a}, Associative Domain Adaptation (ASSC)~\cite{Haeusser_2017_ICCV}, Deep Reconstruction-Classification Network (DRCN)~\cite{DBLP:conf/eccv/GhifaryKZBL16}, MCD~\cite{saito2018maximum}, Virtual Mixup Training (VMT)~\cite{mao2019virtual}, Virtual Adversarial Domain Adaptation (VADA)~\cite{shu2018a} and Decision-boundary Iterative Refinement Traning (DIRT)~\cite{shu2018a}. VMT, VADA and DIRT are Lipschitz-constraint-based methods. The parameters of their backbone are more than twice those of other models, and thus they perform much better than others when no adaption is adopted. 

Among all four tasks, SRDA$^G$ ranks the third in SVHN$\rightarrow$MNIST. However, due to the source-only model of VMT and DIRT performs much better than that of SRDA$^G$, the improvement bought from SRDA$^G$ is the largest.  SRDA$^G$ ranks the fisrt in MNIST$\rightarrow$USPS and SRDA$^F$ performs the best in USPS$\rightarrow$MNIST. Especially, in the most difficult task, i.e., USPS$\rightarrow$MNIST, our three models are the top three. Only MCD and the Lipschitz-constraint-based models are comparable to the proposed models and other methods are inferior to ours with a large margin. In SYNSIG$\rightarrow$GTSRB, our models do not obtain the best results. We reason that it is caused by relatively satisfying results when no adaptation is implemented. Once a model without adaptation has already formed a large margin between different categories, SRDA is hard to detect enough sensitive samples to optimize $G$.
\begin{table*}[htbp]
\centering
\caption{Classification accuracy percentage of VisDA classification experiment. The first row corresponds to the performance if no adaption is implemented. Columns in the middle correspond to different categories and the column on the right represents average accuracy. We evaluate three SRDA models with different methods for adding noise. SRDA* denotes the models that are optimized in an image level. The number behind MCD denotes different hyper-parameters. The results are cited from each study.}
\label{tab:visda}
\begin{tabular}{l|cccccccccccc|r}
\hline\hline
Method&Plane&Bcycl&Bus&Car&Horse&Knife&Mcycl&Person&Plant&Sktbrd&Train&Truck&Mean\\
\hline
Source Only&55.1&53.3 &61.9 &59.1&80.6&17.9&79.7&31.2&81.0&26.5&73.5&8.5&52.4\\
\hline
DAN~\cite{pmlr-v37-long15}&87.1&63.0&76.5&42.0&90.3&42.9&85.9&53.1&49.7&
36.3&{\bf 85.8}&20.7&61.1\\
DANN~\cite{ganin2016domain}&81.9&{\bf 77.7}&82.8&44.3&81.2&29.5&65.1&28.6&
51.9&54.6&82.8&7.8&57.4\\
MCD($2$)~\cite{saito2018maximum}&81.1&55.3&83.6&{\bf 65.7}&87.6&72.7&83.1&73.9&85.3&
47.7&73.2&27.1&69.7\\
MCD($3$)~\cite{saito2018maximum}&90.3&49.3&82.1&62.9&{\bf 91.8}&69.4&83.8&72.8&79.8&
53.3&81.5&29.7&70.6\\
MCD($4$)~\cite{saito2018maximum}&87.0&60.9&{\bf 83.7}&64.0&88.9&79.6&84.7&76.9&88.6&
40.3&83.0&25.8&71.9\\
GPDA~\cite{8953535}&83.0&74.3&80.4&66.0&87.6&75.3&83.8&73.1&{\bf 90.1}&
{\bf 57.3}&80.2&{\bf 37.9}&73.3\\
\hline
VMT~\cite{mao2019virtual}&0.0&0.0&100.0&0.0&0.0&0.0&0.0&0.0&
0.0&0.0&0.0&0.0&8.5\\
VADA~\cite{shu2018a}&93.0&0.0&0.0&0.0&0.0&0.0&0.0&5.0&
0.0&0.0&0.0&0.0&12.7\\
DIRT~\cite{shu2018a}&45.1&0.0&19.7&9.3&0.1&0.0&37.8&0.0&
0.2&7.5&17.1&13.5&13.9\\
\hline
SRDA$^F$&90.1&67.0&82.3&56.0&84.8&{\bf 88.2}&90.3&77.0&
82.5&26.8&85.0&16.2&71.1\\
SRDA$^{F*}$&0.8&0.9&8.8&3.0&1.0&12.9&69.5&0.1&
2.5&0.7&26.8&1.4&11.9\\
SRDA$^V$&89.4&43.5&81.2&60.2&81.1&57.6&{\bf 93.7}&76.6&
81.8&41.3&79.6&22.0&69.5\\
SRDA$^{V*}$&2.6&1.4&2.1&1.6&4.1&23.3&48.9&0.7&
21.8&1.4&3.7&1.3&9.8\\
SRDA$^G$&{\bf 90.9}&74.8&81.9&59.1&87.5&77.3&89.9&{\bf 79.4}&
85.3&40.6&85.1&21.6&{\bf 73.5}\\
SRDA$^{G*}$&40.2&40.2&55.9&62.9&60.5&75.9&83.0&61.7&
73.0&23.2&80.8&5.7&58.0\\
\hline\hline
\end{tabular}
\end{table*}
\subsubsection{VisDA Classification}

Results of the VisDA classification experiment are shown in Table~\ref{tab:visda}. We compare SRDA$^F$, SRDA$^V$ and SRDA$^G$ with several typical methods, such as DAN~\cite{pmlr-v37-long15}, DANN~\cite{ganin2016domain}, MCD~\cite{saito2018maximum}, GPDA~\cite{8953535}, VMT~\cite{mao2019virtual}, VADA~\cite{shu2018a} and DIRT~\cite{shu2018a}. Because VMT, VADA and DIRT are not reported on VisDA in their original papers, we modify their codes and test them on VisDA.

SRDA and GPDA~\cite{8953535} achieve much better accuracy than other methods. Moreover, SRDA$^G$ ranks the first among all the models and SRDA$^F$ obtains comparable accuracy with MCD~\cite{saito2018maximum}. Besides SRDA, other Lipschitz-constraint-based methods perform poorly on VisDA, because the large size of images in VisDA causes a large $d$ that leads to a loose error bound. In detail, SRDA$^G$ achieves the best results in class plane and person, SRDA$^V$ achieves the best result in class motorcycle and SRDA$^F$ gets the best result in class knife. An interesting phenomenon is that three models of SRDA perform diversely among these categories. For example, in class knife, SRDA$^F$ performs much better than the others and SRDA$^V$ ranks the first in class motorcycle. Overall, SRDA$^G$ performs the best. Different performance of three SRDA models reflects the importance of detecting sensitive samples. A well-defined method that can seek more sensitive samples and a metric that can illustrate smoothness of samples precisely are hopeful to further promote the proposed method.

\subsubsection{Office-31 classification}
Results of the Office-31 classification experiment are shown in Table~\ref{tab:office}. We compare SRDA$^F$ and SRDA$^G$ with several UDA methods, such as Geodesic Flow Kernel for unsupervised domain adaptation (GFK)~\cite{gong2012geodesic}, Transfer Component Analysis (TCA)~\cite{pan2010domain}, Residual Transfer Network (RTN)~\cite{long2016unsupervised}, DAN~\cite{pmlr-v37-long15} and DANN~\cite{ganin2016domain}. 

Overall, SRDA$^G$ is comparable to RTN and performs worse than DANN. In particular, SRDA$^F$ performs poorly in D$\rightarrow$A and W$\rightarrow$A whereas in A$\rightarrow$D, A$\rightarrow$W, its performance is acceptable. Considering the extremely small scale of D and W, the results are reasonable. Because of limited samples, SRDA is hard to detect enough sensitive samples to construct a robust decision boundary. In detail, to form a smooth representation space, SRDA seeks sensitive samples and enforces them to hold consistent outputs within their neighbors. An adequate number of such neighbors construct a smooth feature space. Without enough sensitive samples, the local-Lipschitz-constraint is hard to satisfy. This experiment confirms our assumption that a small sample amount of the target domain leads to a loose error bound.

\begin{table}[htbp]
\centering
\caption{Classification accuracy percentage of Office-31 classification experiment among all four tasks. The first row corresponds to the performance if no adaption is implemented. We evaluate two SRDA models with different methods for adding noise. The results are cited from each study.}
\setlength{\tabcolsep}{3.5mm}{
\begin{tabular}{l|ccccc}
\hline\hline
\multirow{3}{*}{Method}& A &A & D & W& \\
&$\rightarrow$ & $\rightarrow$ & $\rightarrow$ & $\rightarrow$& AVG\\
&D & W &A &A &\\
\hline
Source Only& 68.9 & 68.4 & 62.5 & 60.7 & 65.2\\
\hline
GFK~\cite{gong2012geodesic}& 74.5 & 72.8 & 63.4 & 61.0 & 67.9 \\
TCA~\cite{pan2010domain}& 74.1 & 72.7 & 61.7 & 60.9 & 67.4 \\
DAN~\cite{pmlr-v37-long15}& 78.6 & 80.5 & 63.6 & 62.8 & 71.4\\
RTN~\cite{long2016unsupervised}& 77.5 & 84.5 & 66.2 & 64.8 & 73.3 \\
DANN~\cite{ganin2016domain}& 79.7 & 82.0 & {\bf 68.2} & {\bf 67.4} & {\bf 74.3} \\
\hline
SRDA$^F$&{\bf 82.5} & {\bf 84.7} & 62.5 & 61.0 & 72.7\\
SRDA$^G$& 78.8 & 83.2 & 67.3& 64.8 & 73.5\\
\hline\hline
\end{tabular}}
\label{tab:office}
\end{table}

\subsection{Discussions}

\subsubsection{Image Level versus Feature Space}

In our optimization strategy, LSD is minimized in feature space instead of an image level. This modification is introduced  because the dimension $d$ of an image greatly impact the error bound of a UDA problem. If the optimization is conducted in an image level, $d$ would be much larger than in feature space and a loose error bound is formed. To verify it, we assess SRDA that optimizes in an image level on both digits and VisDA datasets. In detail, we adopt (\ref{eq6}) and (\ref{eq9}) as our optimization goals. Three SRDA models that generate noise from FSGM, VAT and a Gaussian distribution are denoted as SRDA$^{F*}$, SRDA$^{V*}$ and SRDA$^{G*}$, respectively. 

\begin{figure*}[h]
  \centering
  \subfigure[SVHN$\rightarrow$MNIST]{
    \label{fig:svhnmnist} 
    \includegraphics[width=0.38\columnwidth]{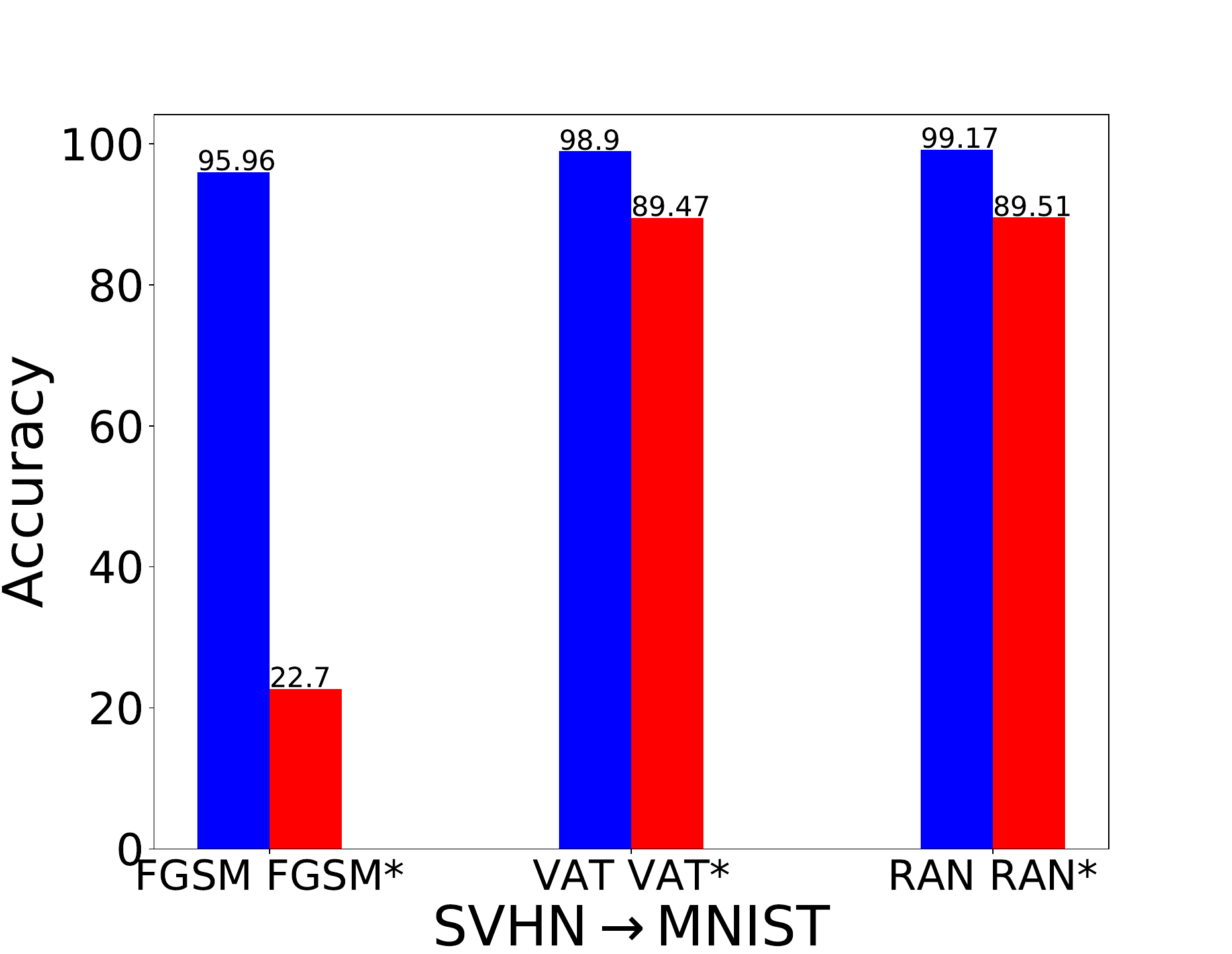}}
  \subfigure[SYNSIG$\rightarrow$GTSRB]{
    \label{fig:synthgtsrb} 
    \includegraphics[width=0.38\columnwidth]{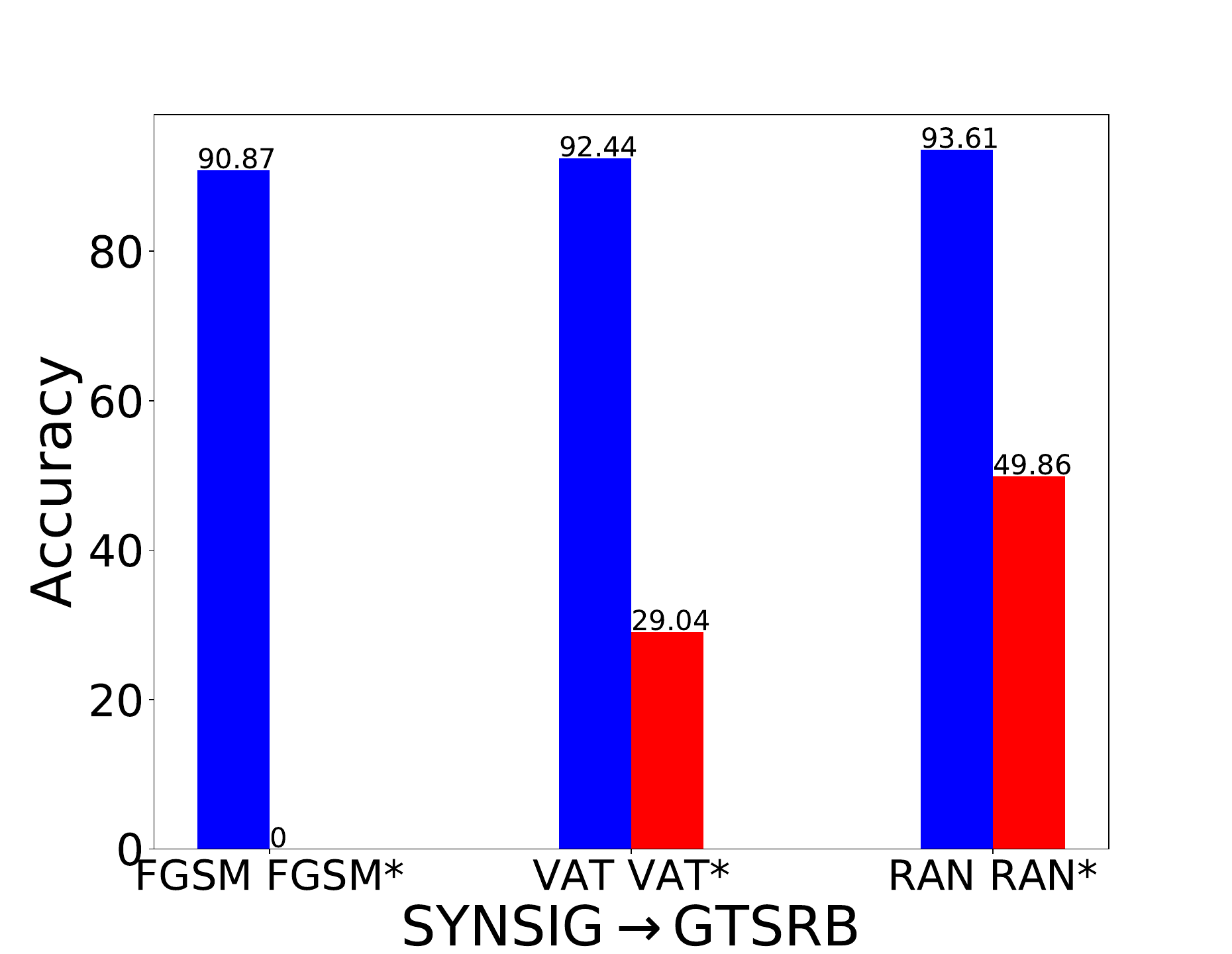}}
  \subfigure[MNIST$\rightarrow$USPS]{
    \label{fig:mnistusps} 
    \includegraphics[width=0.38\columnwidth]{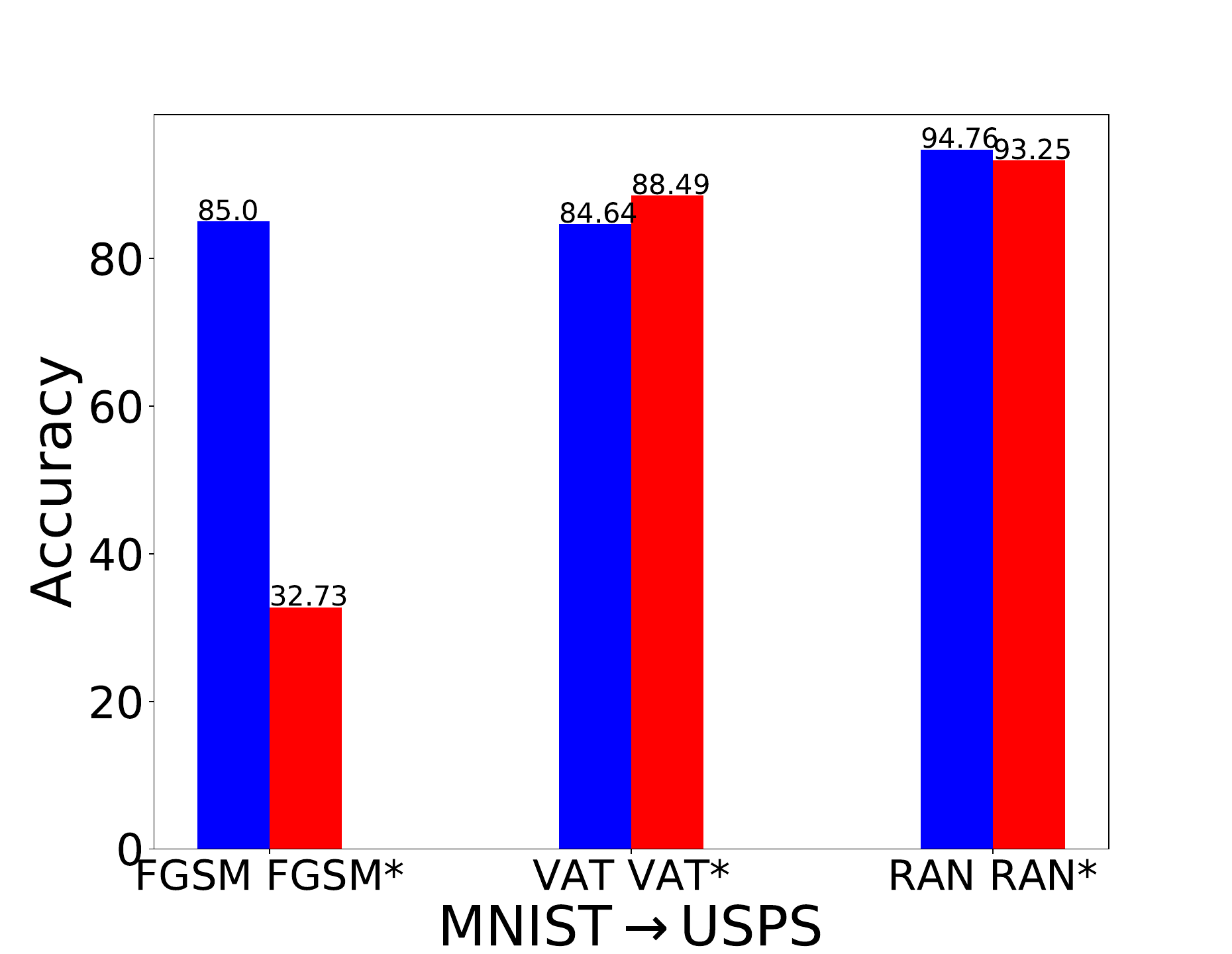}}
	\subfigure[USPS$\rightarrow$MNIST]{
    \label{fig:uspsmnist} 
    \includegraphics[width=0.38\columnwidth]{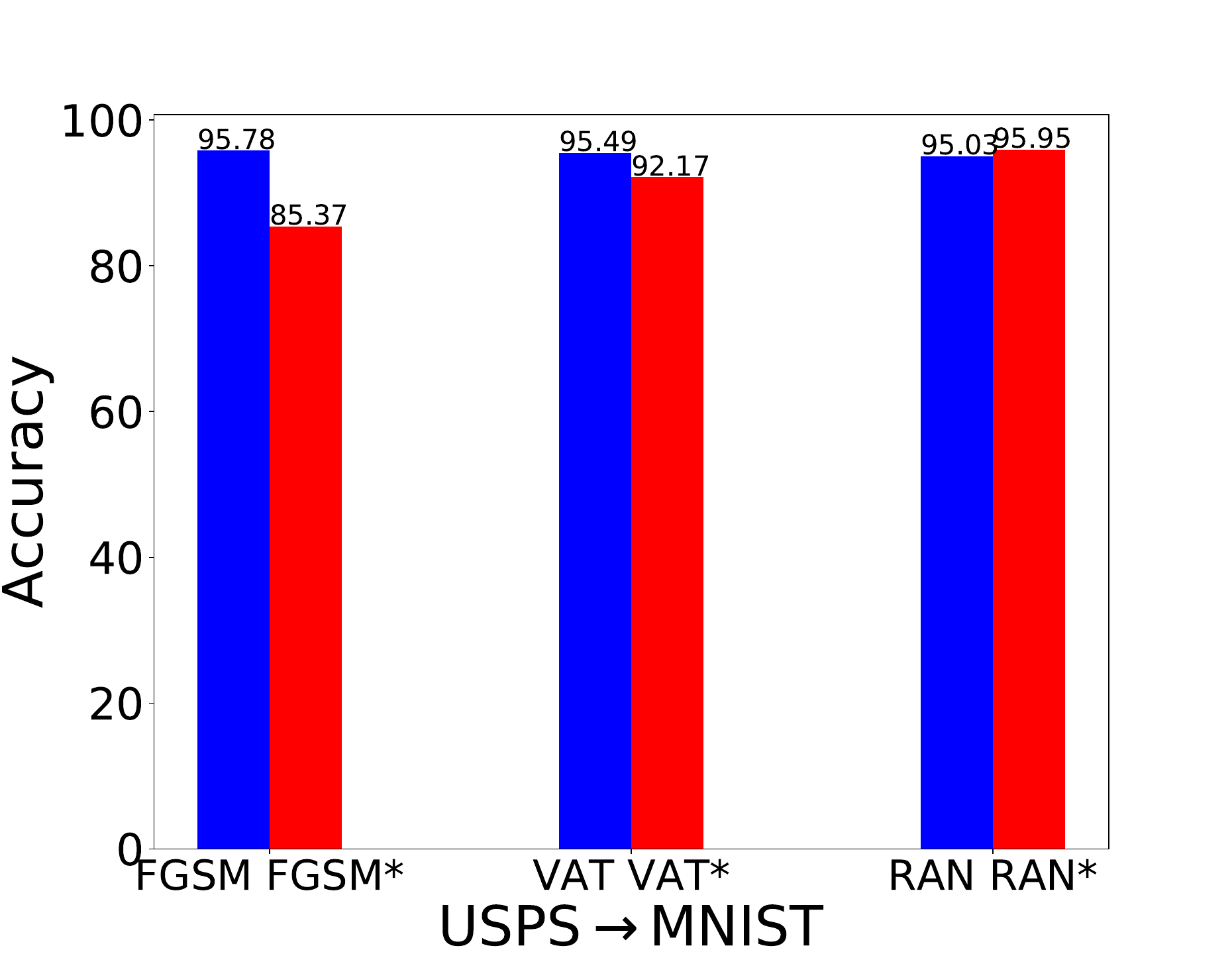}}
\subfigure[VisDA]{
    \label{fig:VisDA} 
    \includegraphics[width=0.38\columnwidth]{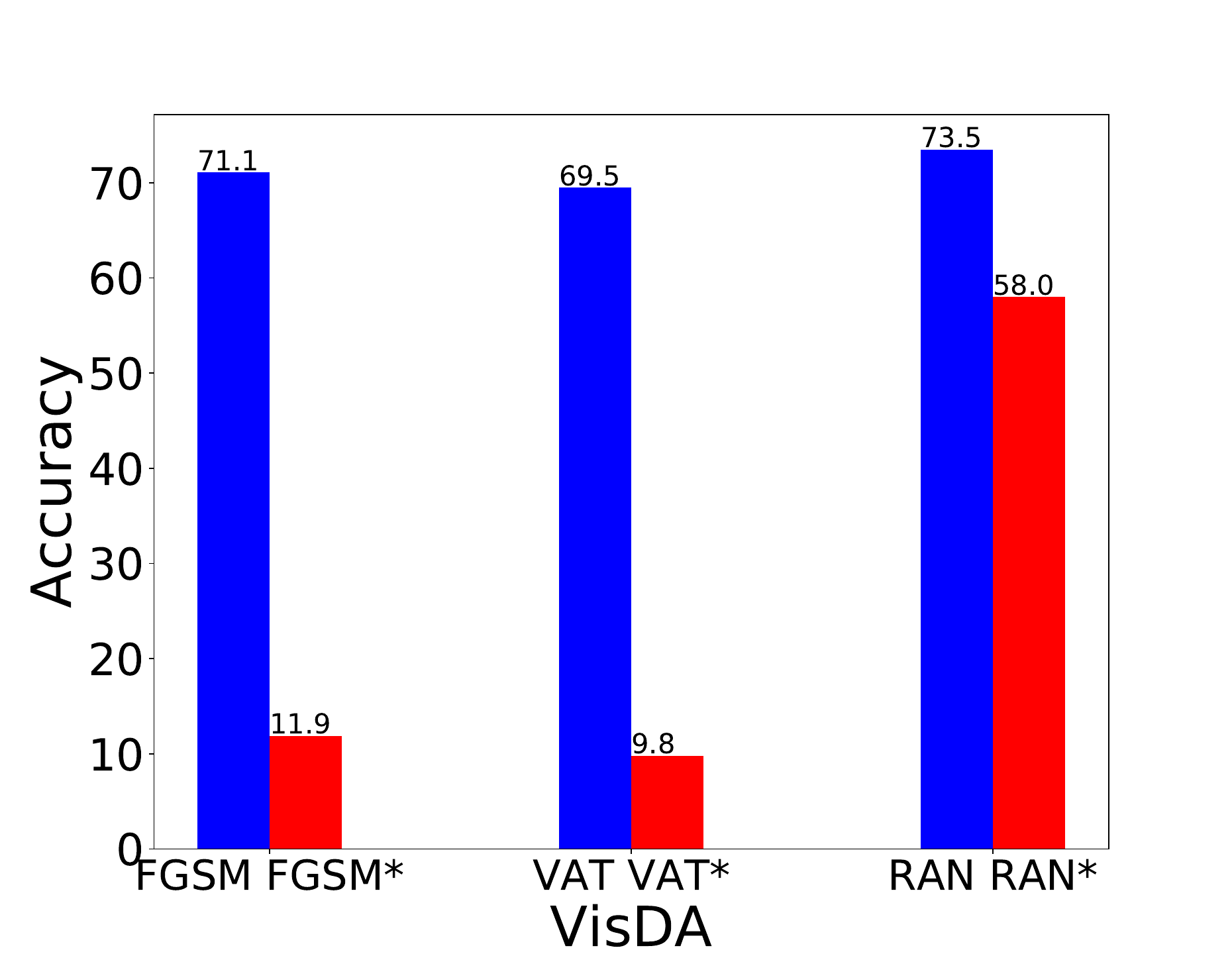}}  
\caption{The performance of SRDA and SRDA* on digits and VisDA datasets. Blue bars denote SRDA while red ones denote SRDA*. FGSM, VAT and RAN denote SRDA$^F$, SRDA$^V$ and SRDA$^G$, respectively. FGSM*, VAT* and RAN* denote SRDA$^{F*}$, SRDA$^{V*}$ and SRDA$^{G*}$, respectively.}
  \label{fig:imagefeature} 
\end{figure*}

Results are shown in Fig.~\ref{fig:imagefeature} and Table~\ref{tab:visda}. Except that SRDA$^{G*}$ and SRDA$^{V*}$ perform slightly better than SRDA$^G$ and SRDA$^V$ in USPS$\rightarrow$MNIST and MNIST$\rightarrow$USPS, respectively, all models optimized in feature space obtain much better performance. Particularly, SRDA$^{F*}$ is sensitive to $d$ where its performance drops a lot in almost all tasks and it even cannot converge in SYNSIG$\rightarrow$GTSRB. On VisDA, SRDA$^{F*}$ and SRDA$^{V*}$ perform like random guessing, and the accuracy of SRDA$^{G*}$ decreases 15.5\%. 

Another evidence is that previous Lipschitz-constraint-based methods, i.e., VMT, VADA and DIRT, perform poorly on VisDA. These methods satisfy the local Lipschitz constraint at an image level. VMT converges to a trivial solution where all images are classified to a category. The accuracy of VADA and DIRT is less than 15\%. On the contrary, they perform excellently on Digits whose images are small-scale.

The experimental results demonstrate that satisfying a Lipschitz constraint in feature space is convenient to apply Lipschitz-based UDA algorithms to large-scale datasets. Large-scale images in VisDA dataset aggravate the influence of $d$ because the difference of $d$ between an image level and feature space is more obvious. Therefore, we conclude that optimization in feature space is necessary to reduce the value of $d$. Considering that previous Lipschitz-based methods~\cite{shu2018a,mao2019virtual} are all conducted on small-scale datasets, the proposed method gives a promising direction to spread them to more scenarios. In addition, the isotropic method is more robust than the anisotropic one when an SRDA model is optimized in an image level. We speculate that all-direction exploration avoids instability.

\begin{figure*}[h]
  \centering
  \subfigure[SRDA$^F$]{
    \label{fig:visda_fgsm} 
    \includegraphics[width=0.31\columnwidth]{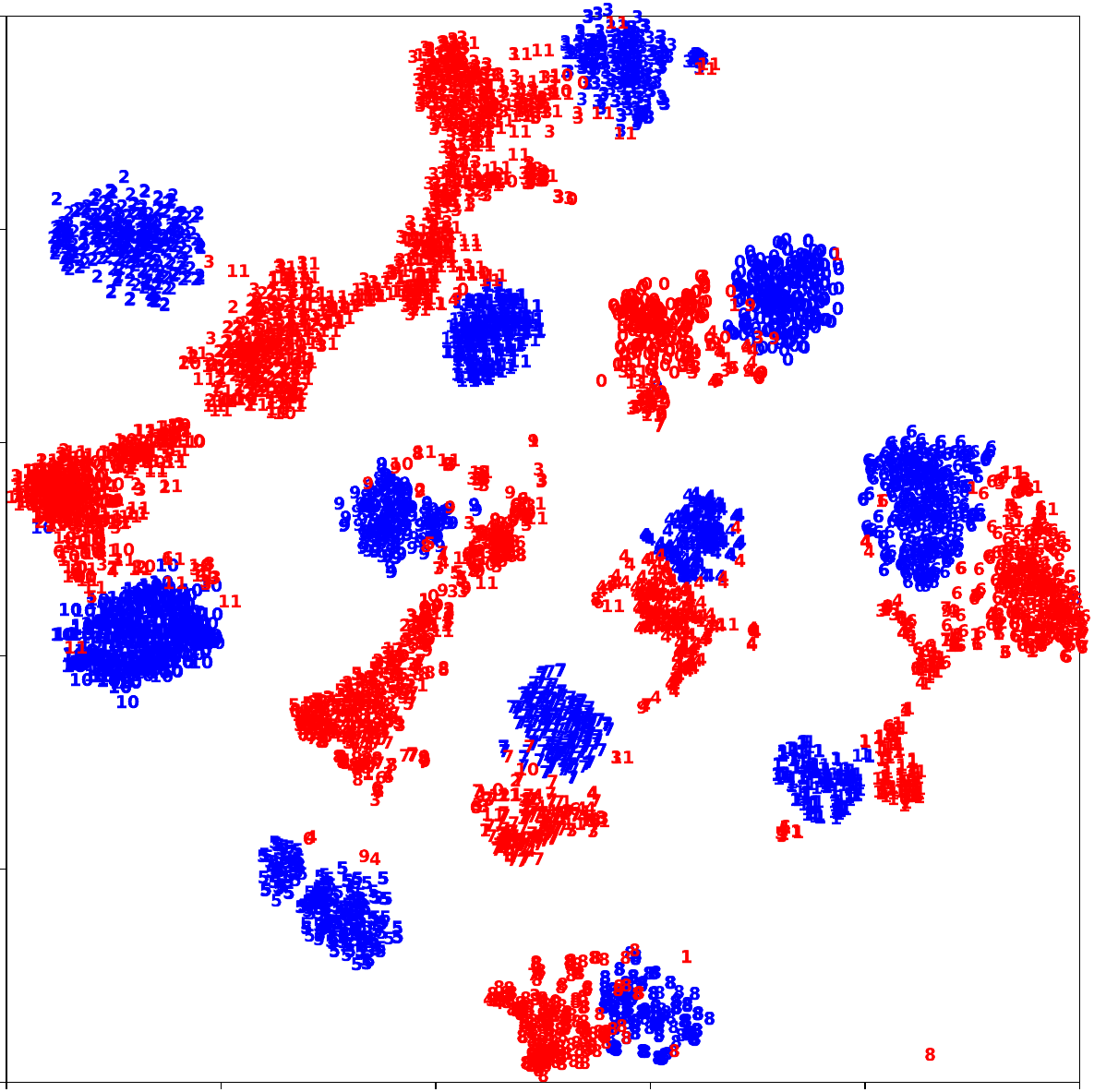}}
  \subfigure[SRDA$^V$]{
    \label{fig:visda_vat} 
    \includegraphics[width=0.31\columnwidth]{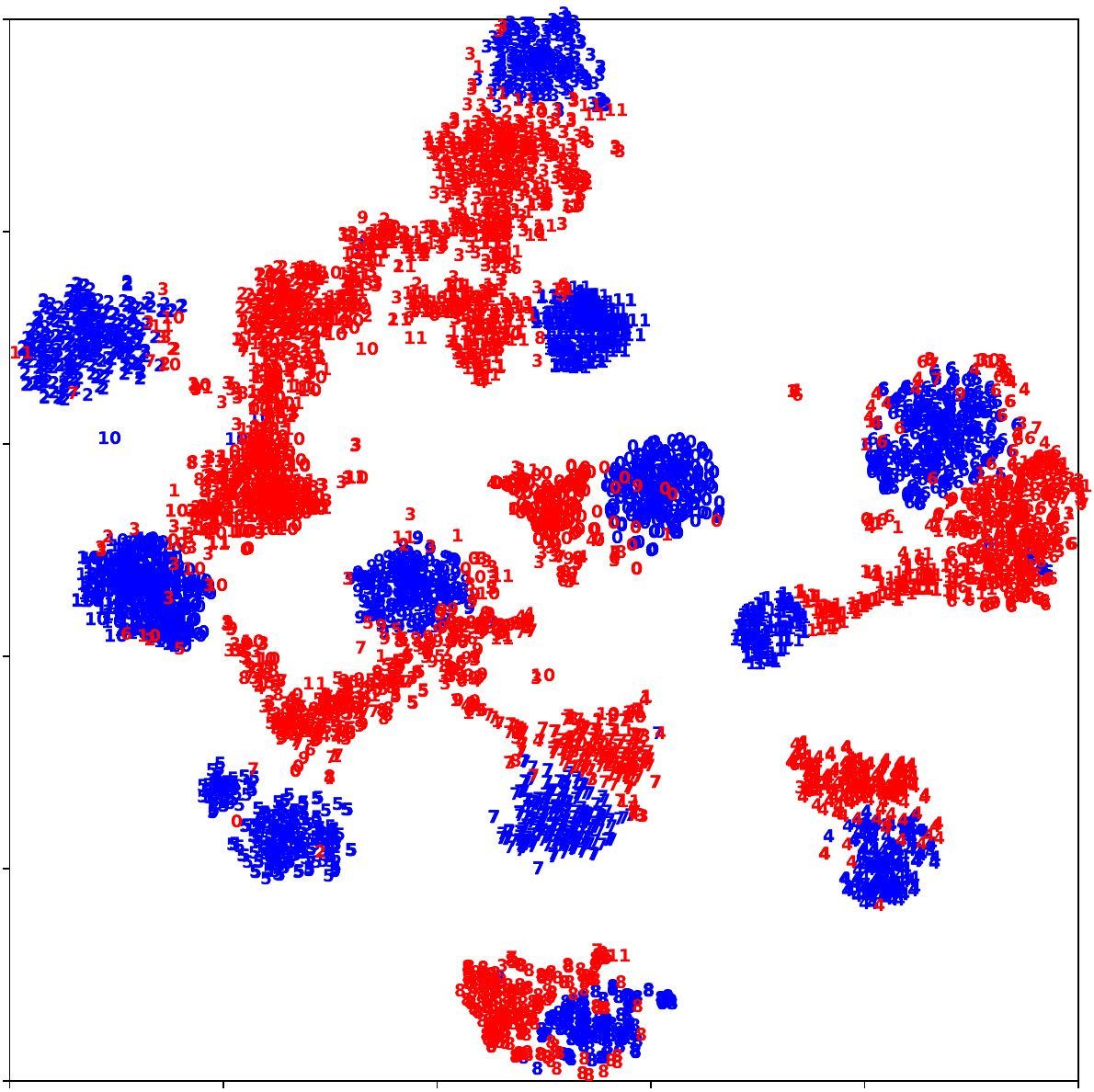}}
	\subfigure[SRDA$^F$]{
    \label{fig:visda_random} 
    \includegraphics[width=0.31\columnwidth]{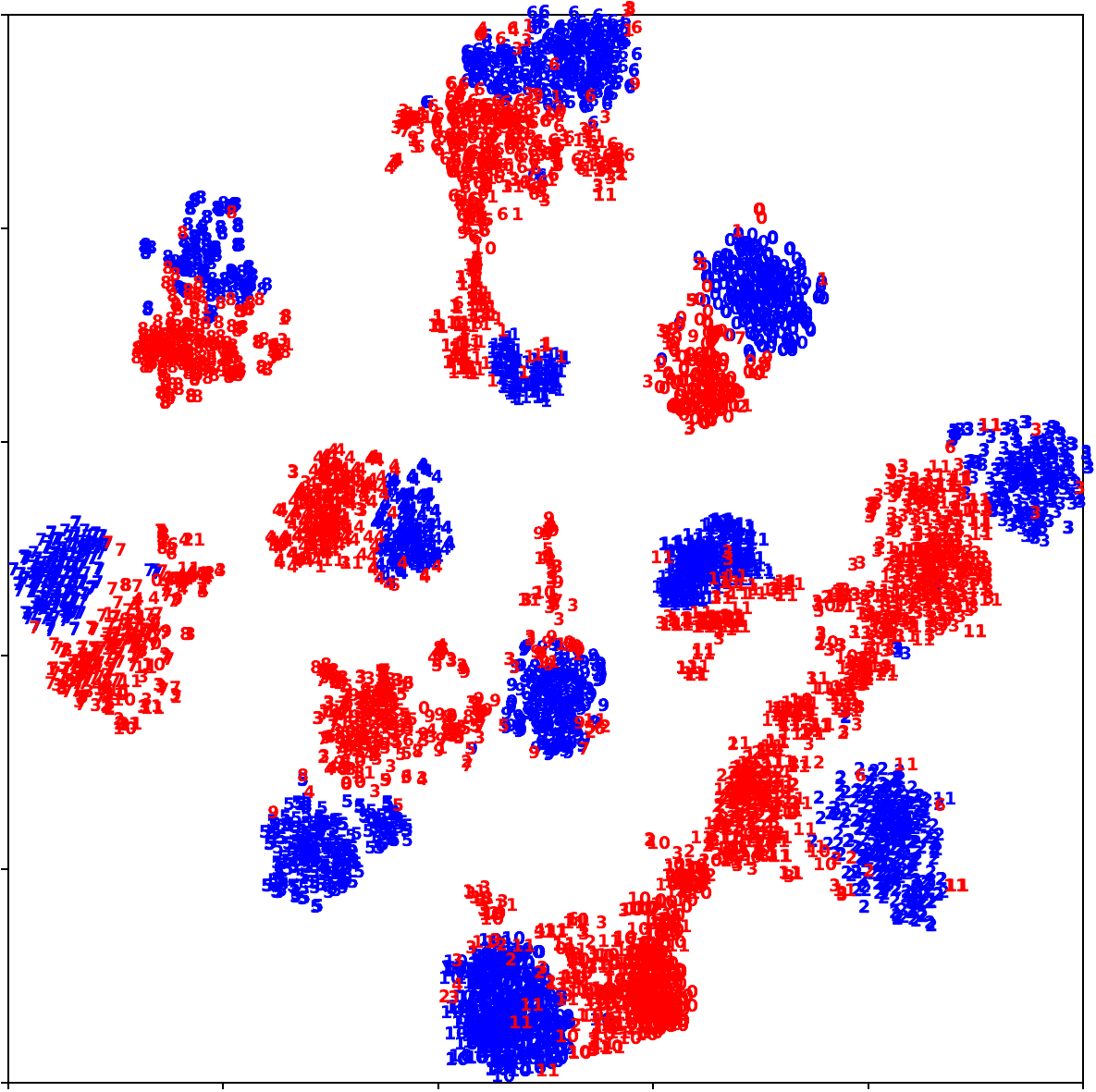}} 
\subfigure[SRDA$^F$]{
    \label{fig:digits_fgsm} 
    \includegraphics[width=0.31\columnwidth]{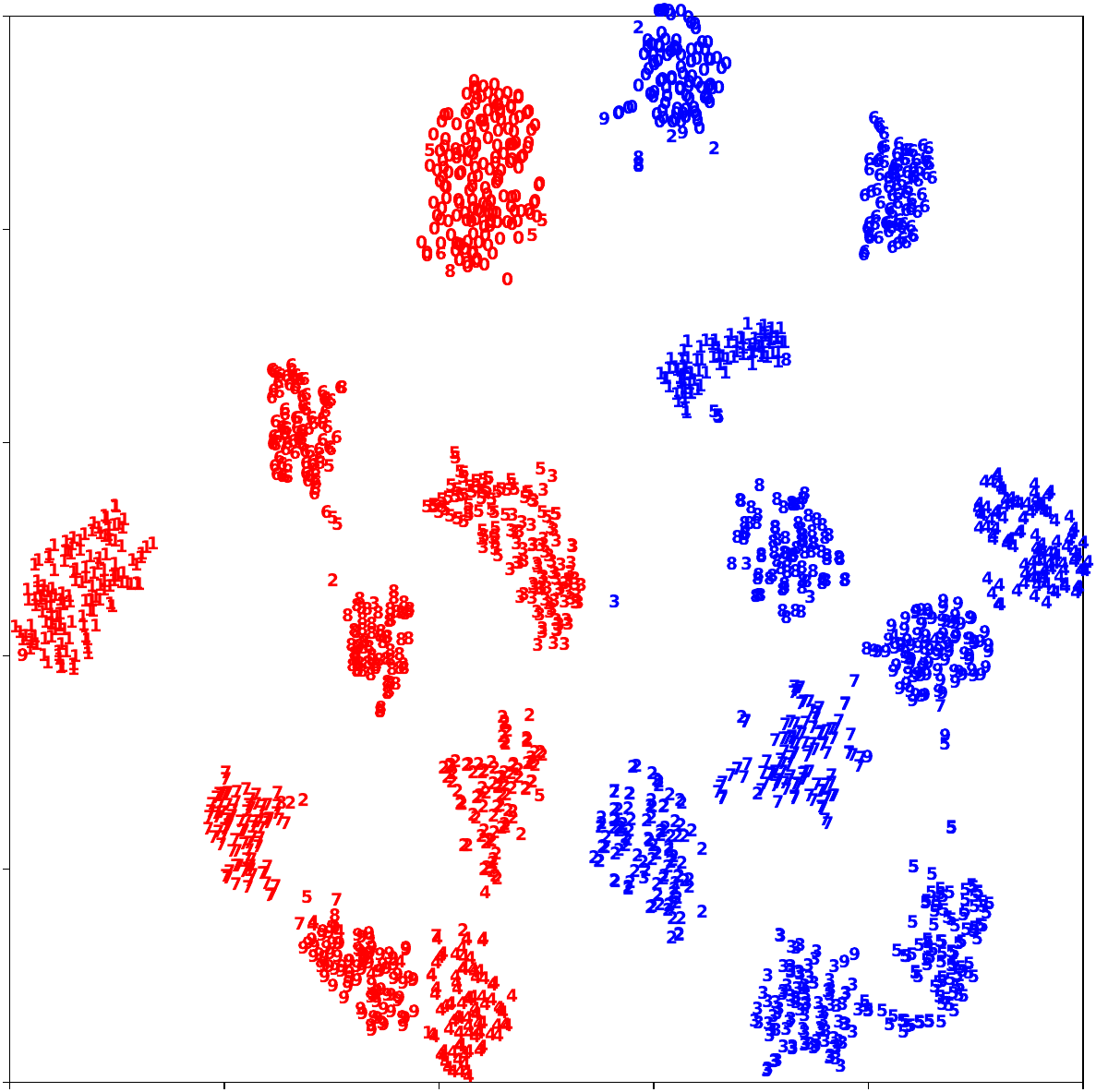}} 
\subfigure[SRDA$^V$]{
    \label{fig:digits_vat} 
    \includegraphics[width=0.31\columnwidth]{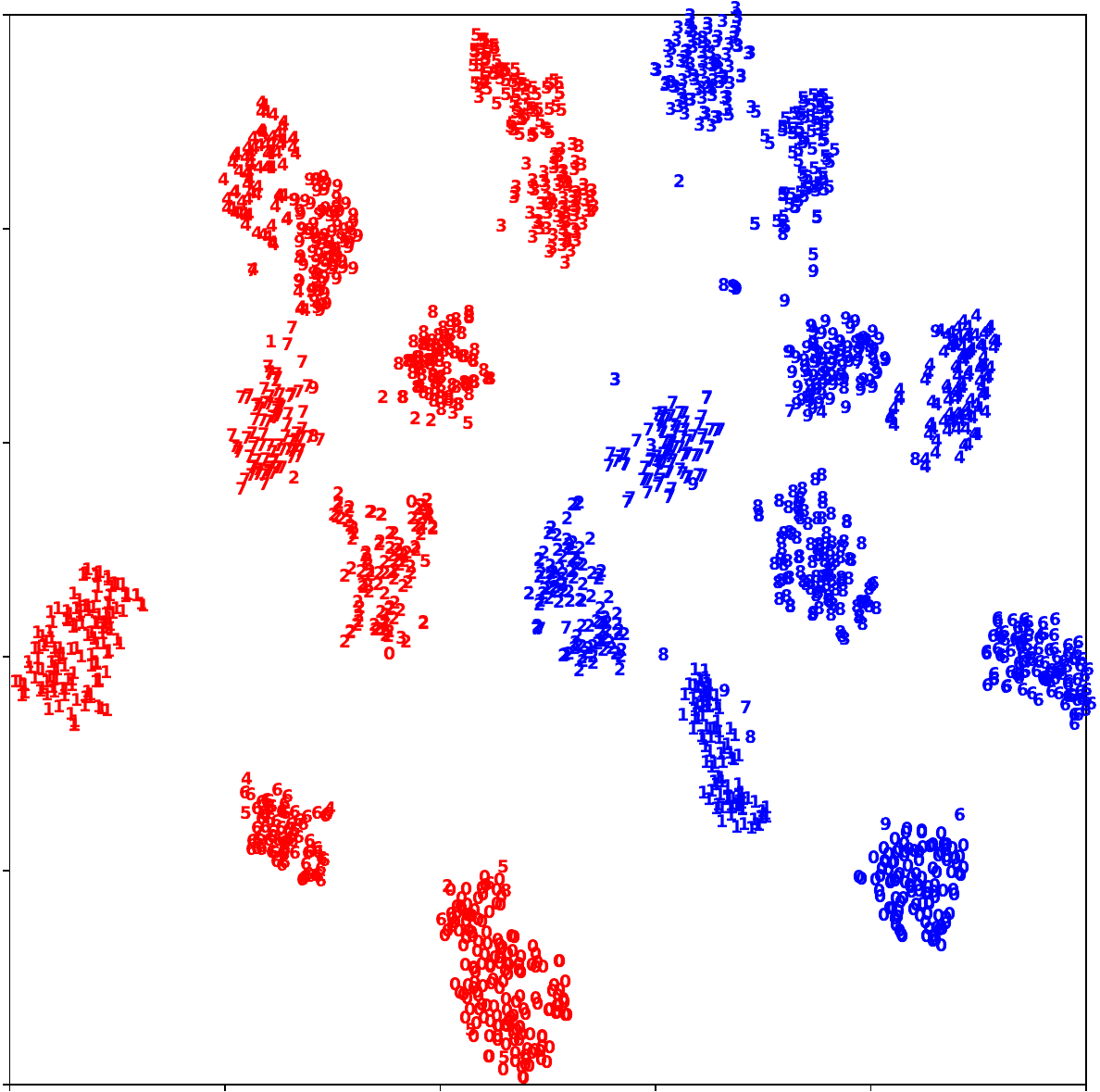}} 
	\subfigure[SRDA$^G$]{
    \label{fig:digits_random} 
    \includegraphics[width=0.31\columnwidth]{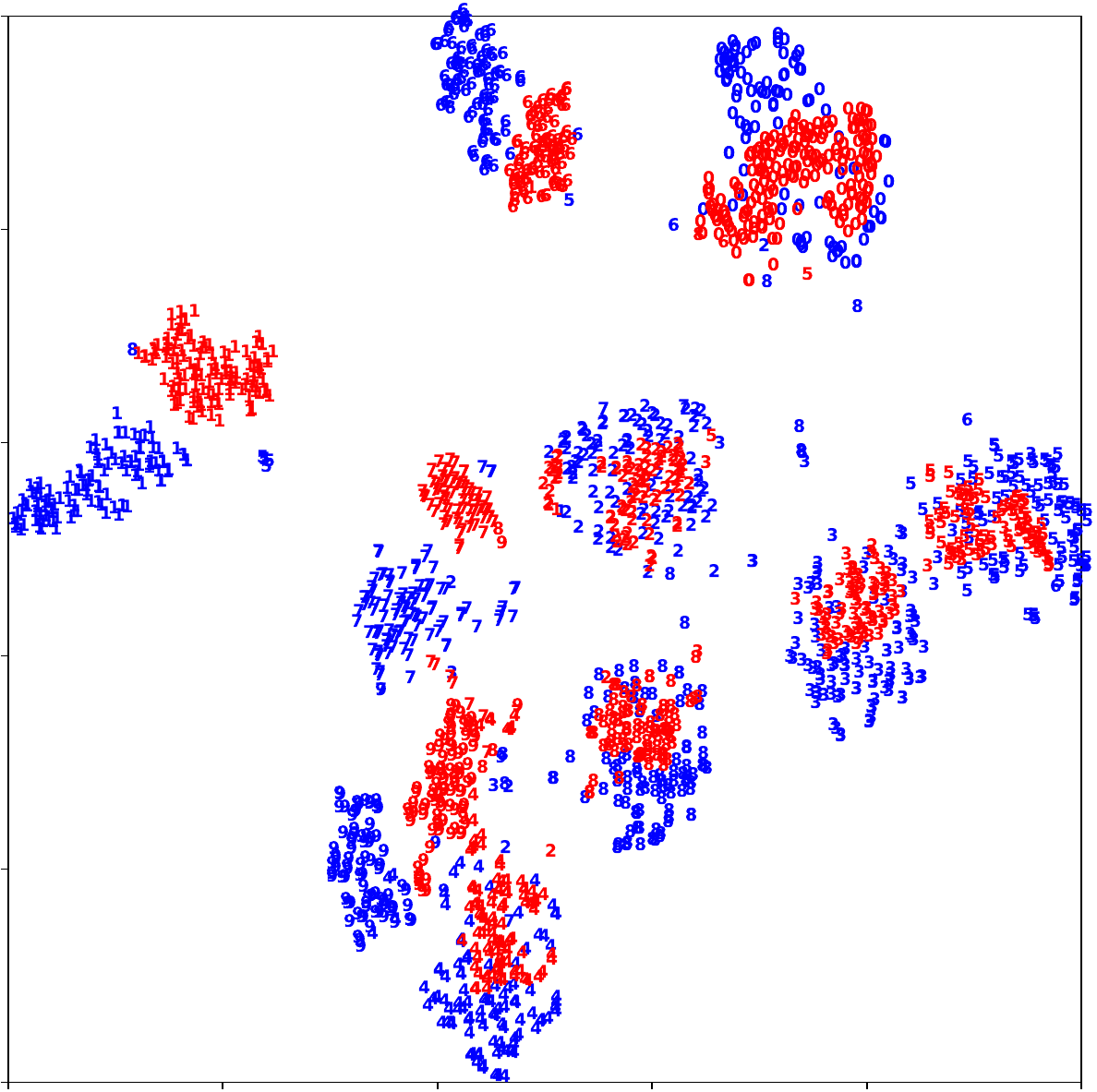}} 
\caption{(a)-(c) T-SNE plots of SRDA for VisDA classification experiment. Blue points denote a source domain while red ones denote a target domain. (d)-(f) T-SNE plots of SRDA for MNIST (blue)$\rightarrow$USPS (red).}
  \label{fig:embedding} 
\end{figure*}

\begin{figure*}[h]
  \centering
  \subfigure[SRDA$^{F*}$]{
    \label{fig:visda_fgsm_all} 
    \includegraphics[width=0.31\columnwidth]{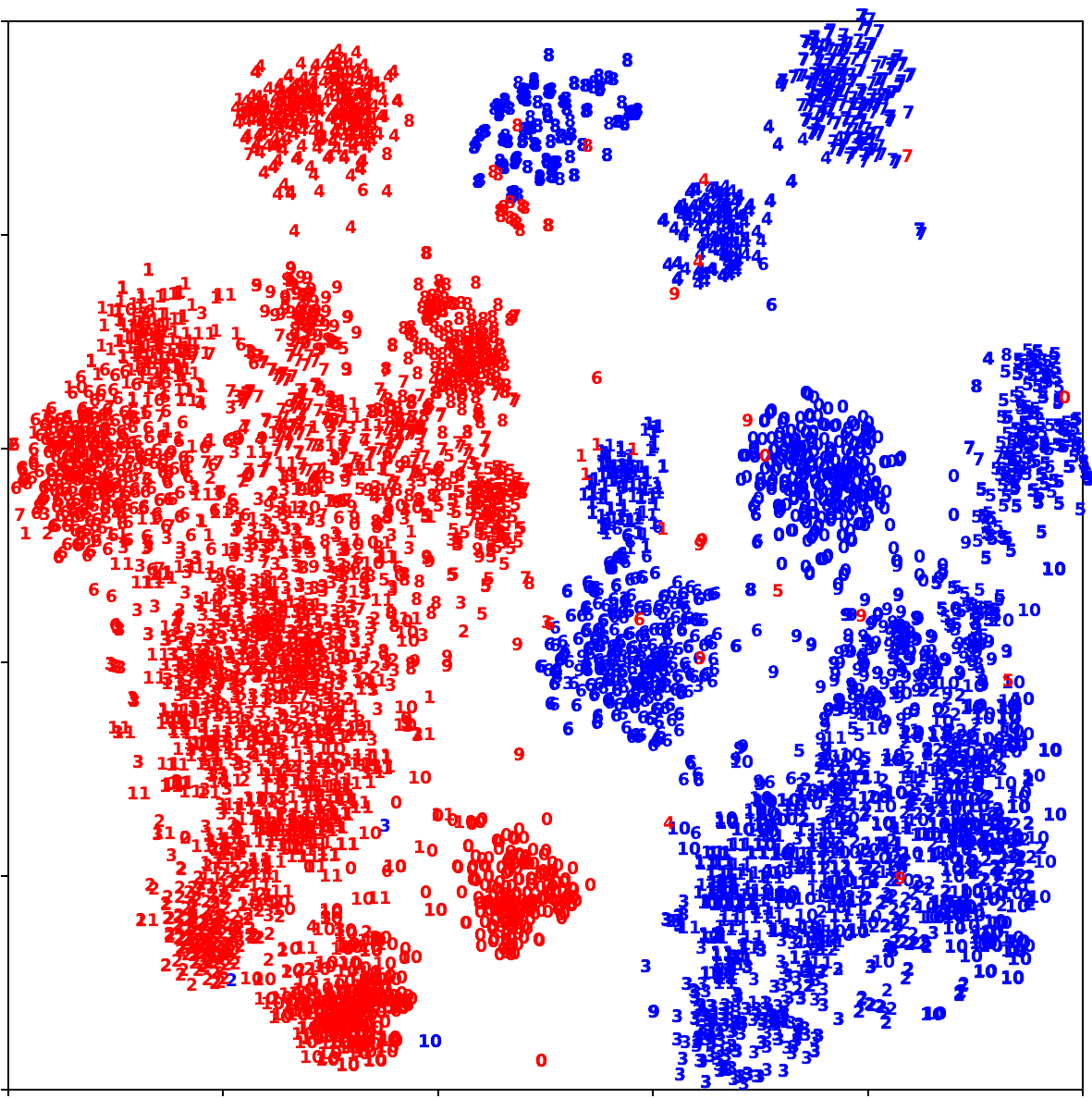}}
  \subfigure[SRDA$^{V*}$]{
    \label{fig:visda_vat_all} 
    \includegraphics[width=0.31\columnwidth]{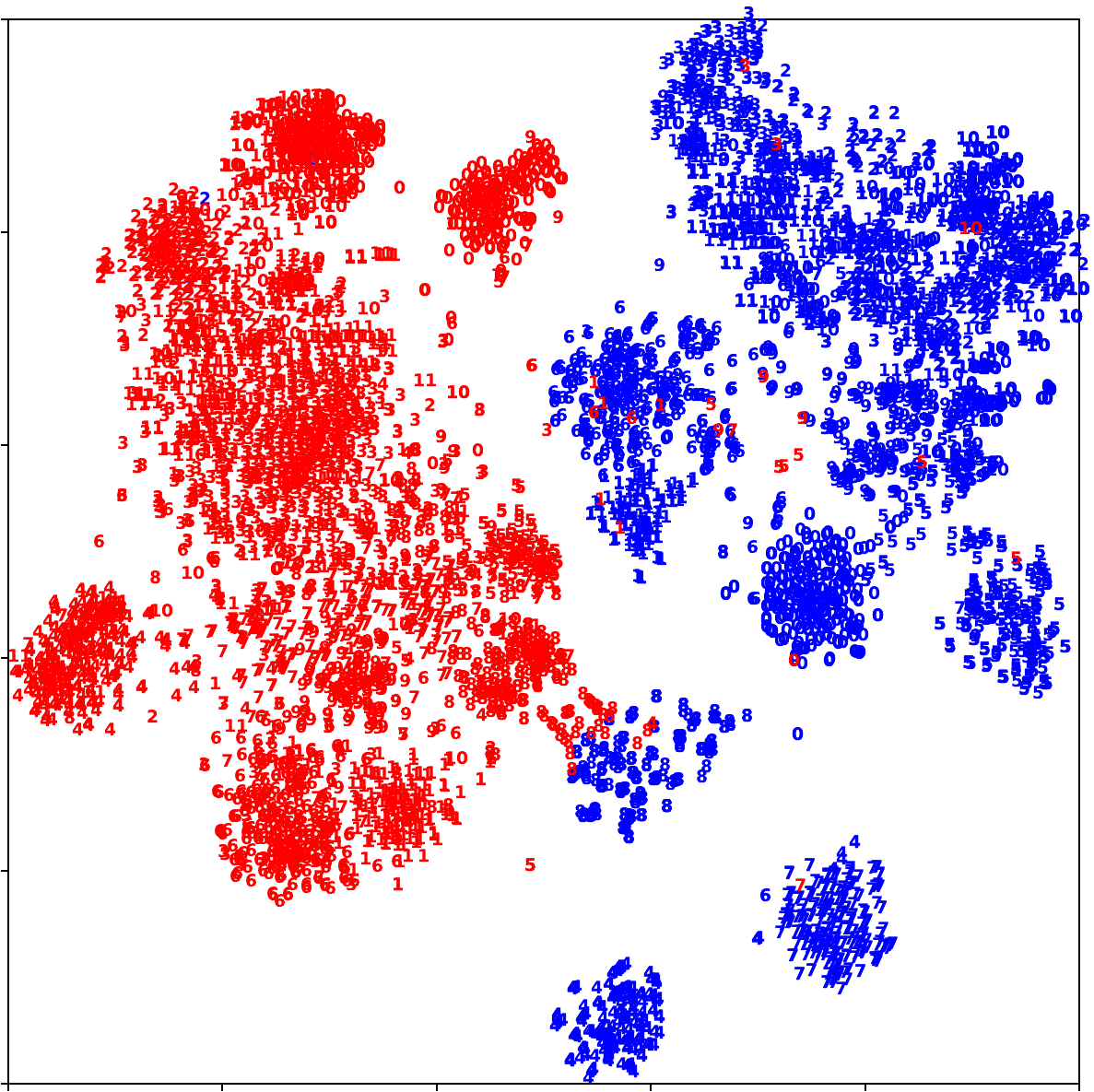}}
	\subfigure[SRDA$^{F*}$]{
    \label{fig:visda_random_all} 
    \includegraphics[width=0.31\columnwidth]{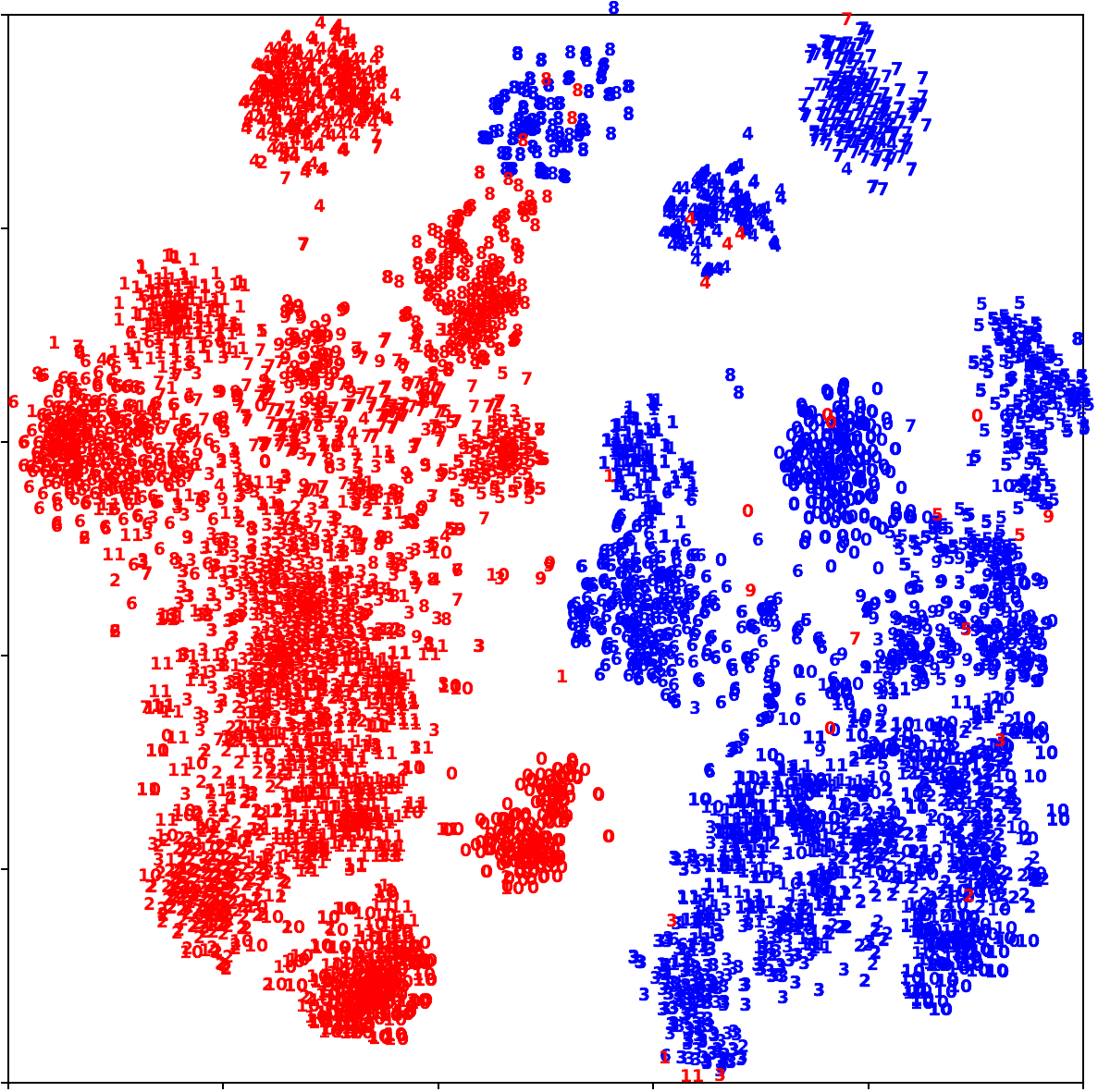}} 
\subfigure[SRDA$^{F*}$]{
    \label{fig:digits_fgsm_all} 
    \includegraphics[width=0.31\columnwidth]{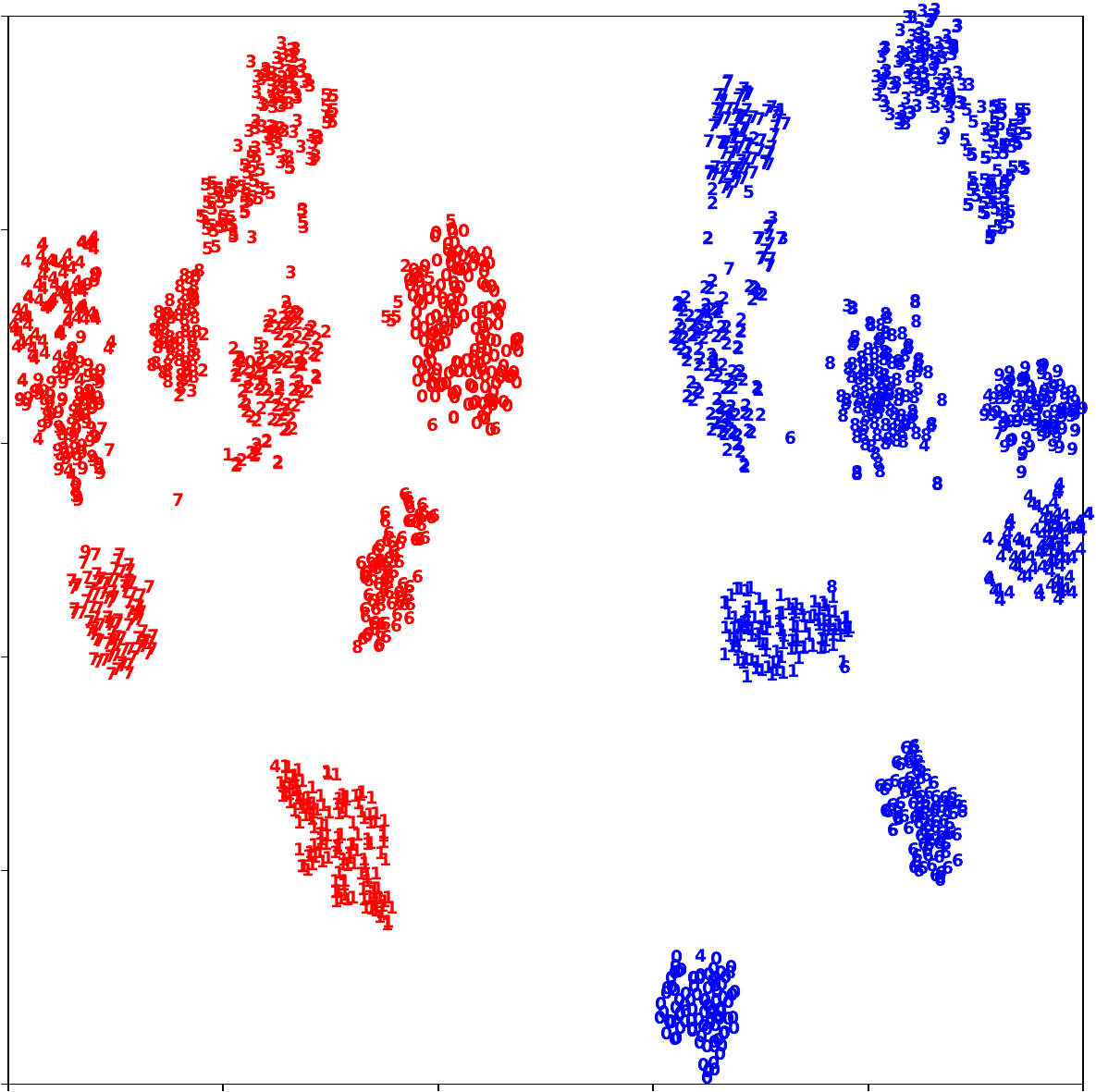}} 
\subfigure[SRDA$^{V*}$]{
    \label{fig:digits_vat_all} 
    \includegraphics[width=0.31\columnwidth]{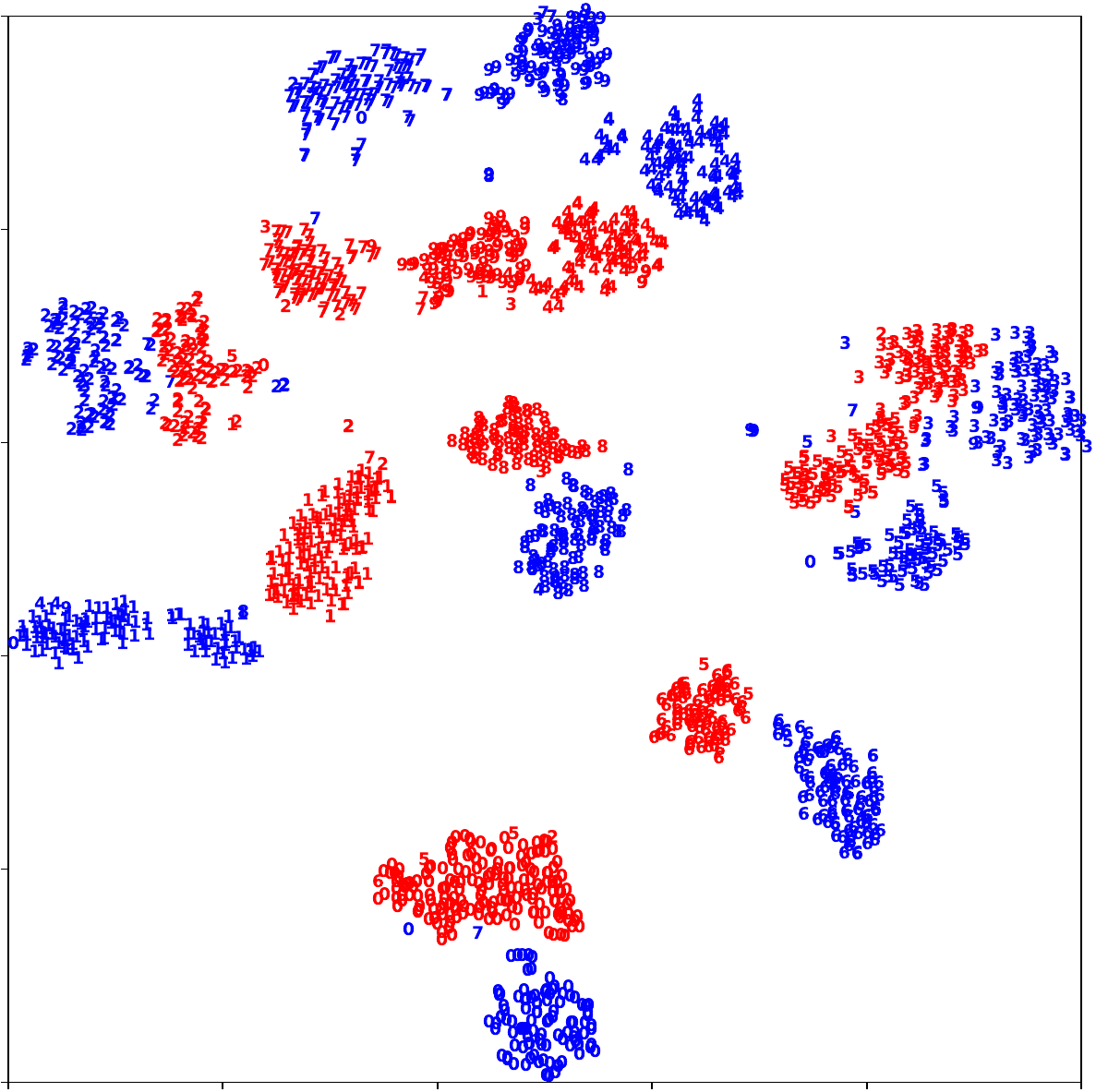}} 
	\subfigure[SRDA$^{G*}$]{
    \label{fig:digits_random_all} 
    \includegraphics[width=0.31\columnwidth]{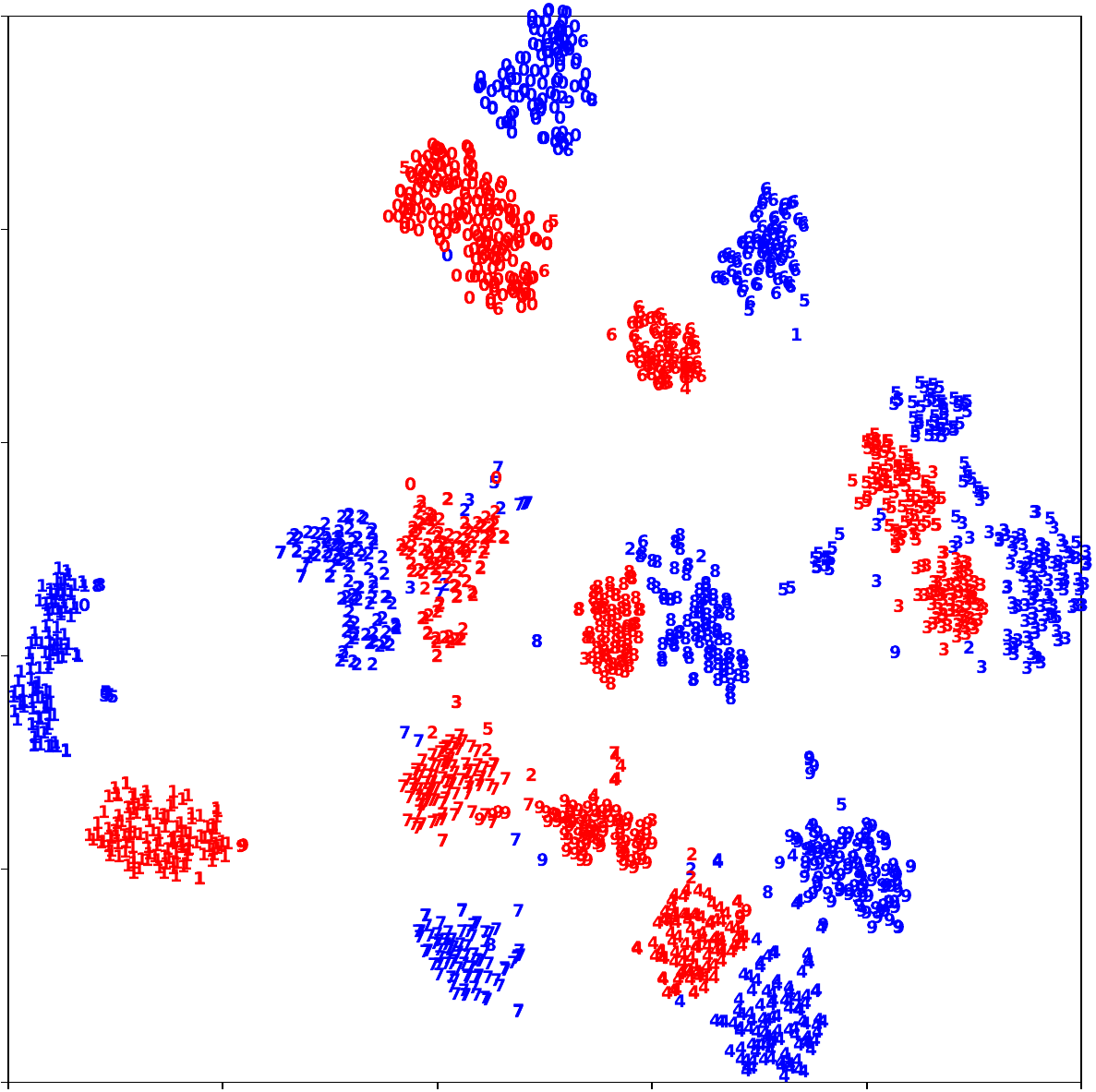}} 
\caption{(a)-(c) T-SNE plots of SRDA* for VisDA classification experiment. Blue points denote a source domain while red ones denote a target domain. (d)-(f) T-SNE plots of SRDA* for MNIST (blue)$\rightarrow$USPS (red).}
  \label{fig:all_embedding} 
\end{figure*}

\subsubsection{Sample Amount of Target Domain}
We conclude that a small sample amount of the target domain would lead to a loose error bound for UDA. In this section, it is verified by comparing the performance of SRDA on VisDA and Office-31. In detail, VisDA includes 55,388 samples in the target domain,  while three domains in Office-31 includes fewer than 3,000 images. To make a fair comparison, we resize all images to 224$\times$224 and use the same network backbone. 

As the results shown in Tables~\ref{tab:visda}-\ref{tab:office}, SRDA$^G$ performs better than DANN and DAN by 16.1\% and 12.4\% on VisDA. However, on Office-31, SRDA$^G$ performs worse than DANN and better than DAN by 2.1\%. Similarly to SRDA$^G$, SRDA$^F$ performs better than DANN and DAN by 13.4\% and 10\% on VisDA. On Office-31, SRDA$^F$ performs worse than DANN and outperforms DAN by 1.3\%. It is obvious that SRDA shows a large advantage over DAN and DANN if the target domain includes a large number of images. On the contrary, if the sample amount of a target domain is small, such as Office-31, the Lipschitz-constraint-based models fail to perform well. The different performance of SRDA on VisDA and Office-31 shows that the sample amount of a target domain is critical for Lipschitz-constraint-based methods' performance.

\subsubsection{Visualization of Representation}

In Figs.~\ref{fig:embedding} and~\ref{fig:all_embedding}, we further analyze the behavior of SRDA and SRDA* by T-SNE embeddings of the feature space where we adopt our optimization strategy. In particular, we visualize embeddings for MNIST$\rightarrow$USPS and VisDA classification. 

In Fig.~\ref{fig:digits_random}, SRDA$^G$ not only separates target samples into ten clusters clearly but also aligns source and target distributions well. However, SRDA$^F$ and SRDA$^V$ only separate target samples into ten clusters as shown in Figs.~\ref{fig:digits_fgsm} and \ref{fig:digits_vat}. It explains why SRDA$^G$ outperforms the other two models in MNIST$\rightarrow$USPS. An alignment of different domains enhances the performance of a target domain. In Figs.~\ref{fig:visda_fgsm}-(c), although target samples are separated into clusters and different domains align well, there exists adhesion among different clusters. Such adhesion impedes a classifier trained on a source domain to classify target samples as their correct categories. This means that a large margin among these clusters is not built. We assume that some sensitive samples belonging to different categories from their neighbors are forced to hold consistent outputs such that the adhesion among different clusters is formed.

In Figs.~\ref{fig:digits_vat_all} and \ref{fig:digits_random_all}, SRDA$^{G*}$ and SRDA$^{V*}$ separate target samples clearly and align different domains well. Compared to SRDA$^V$, SRDA$^{V*}$ even matches different domains better. It explains why SRDA$^{V*}$ achieves better accuracy than SRDA$^V$ in MNIST$\rightarrow$USPS. In Fig.~\ref{fig:digits_fgsm_all}, SRDA$^{F*}$ performs like SRDA$^F$. They form clear clusters but fail to match source and target domains. In Figs.~\ref{fig:visda_fgsm_all}-(c), all clusters assemble together and two domains are separated in all three models. A classifier is hard to work well in this situation. This phenomenon demonstrates that holding a Lipschitz constraint in an image level is unfeasible on a large-scale dataset like VisDA.

\subsubsection{Batchsize Analysis}
\label{batchsize_exp}

Besides the dimension of samples and sample amount of a target domain,  we discover that a large batchsize improves SRDA's performance. A large batchsize is recommended in other computer vision problems, such as image generation~\cite{brock2018large} and image classification~\cite{He_2019_CVPR}. However, to our best knowledge, this work is the first one to consider whether a large batchsize can help us solve a UDA problem. To verify how a batchsize influences the performance of SRDA, we evaluate it with $m\in\{32, 16, 8, 4\}$. Except for batchsize, other settings follow the VisDA classification experiment.

Results are shown in Table~\ref{tab:batchsize}. For all the three SRDA models, the average accuracy decreases as their batchsize gets smaller. Especially, when the batchsize is set to $4$, the performance drops rapidly even lower than $30\%$. When the batchsize $\geq8$, the performance drops gradually  as the batchsize decreases and three models maintain their accuracy over $60\%$. The trend shows that the batchsize should be set large enough for optimizing SRDA.
\begin{table}[htbp]
\centering
\caption{Classification accuracy percentage of VisDA classification experiment for different batchsize.}
\label{tab:batchsize}
\setlength{\tabcolsep}{5mm}{
\begin{tabular}{l|cc}
\hline\hline
Method& Batchsize & Average Accuracy \\
\hline
\multirow{4}{*}{SRDA$^F$}& 32& {\bf 71.1}\\
& 16& 69.1\\
& 8& 64.5\\
& 4& 29.5\\
\hline
\multirow{4}{*}{SRDA$^V$}& 32& {\bf 69.5}\\
& 16& 66.5\\
& 8& 62.2\\
& 4& 34.3\\
\hline
\multirow{4}{*}{SRDA$^G$}& 32& {\bf 73.5}\\
& 16& 71.1\\
& 8& 64.2\\
& 4& 51.4\\
\hline\hline
\end{tabular}}
\end{table}

\subsubsection{Sensitivity Analysis}

We add a detailed discussion on $\epsilon$. We test $\epsilon\in\{0.3, 0.4, 0.5, 0.6, 0.7\}$ on the digits datasets. Experimental results indicate that they are not sensitive to $\epsilon$; and thus we do not need to set $\epsilon$ carefully. We can easily set $\epsilon = 0.5$ as a default value. 

\begin{figure}
  \centering
  \subfigure[USPS$\rightarrow$MNIST]{
    \label{fig:uspsmnistsensitivity} 
    \includegraphics[width=0.48\columnwidth]{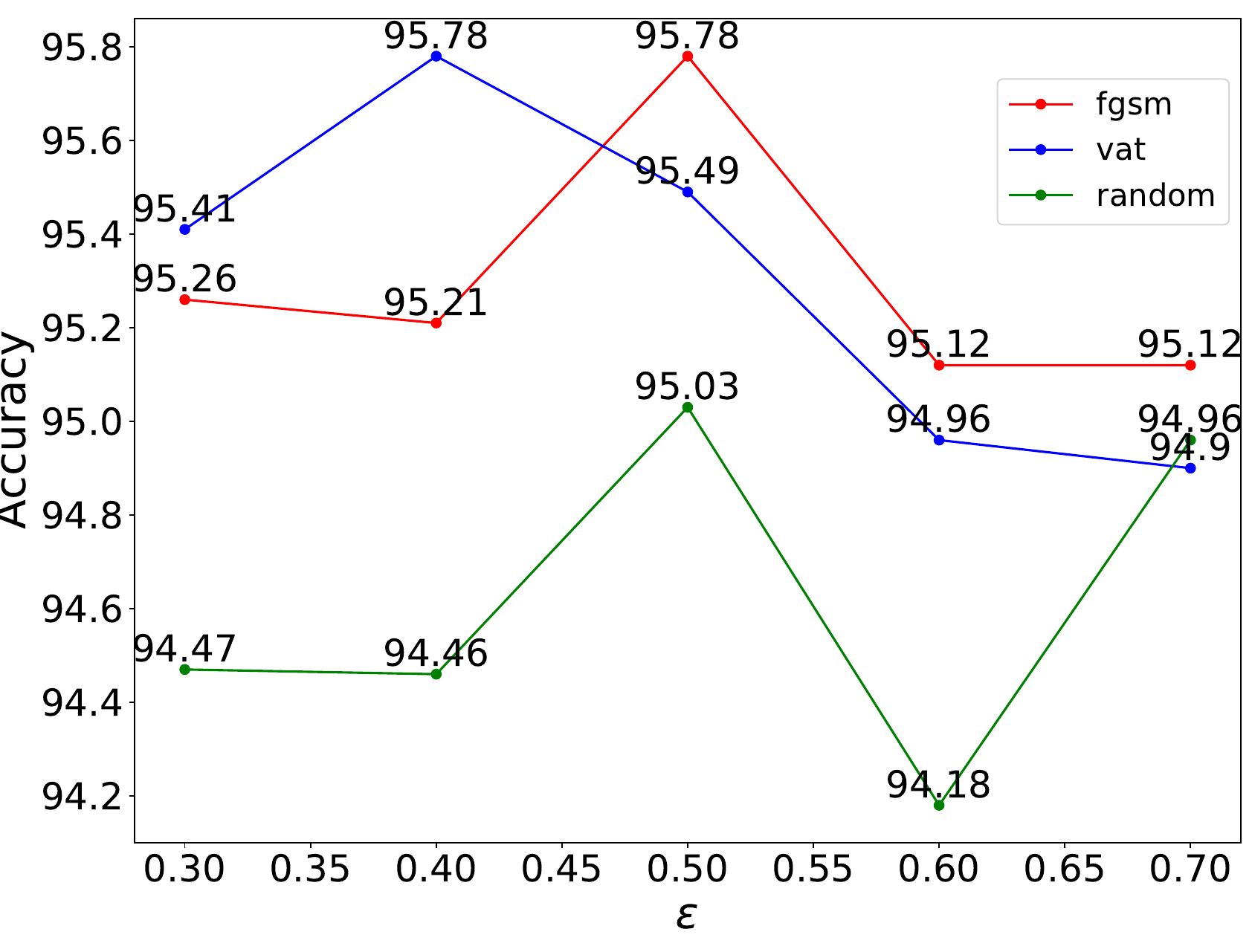}}
  \subfigure[MNIST$\rightarrow$USPS]{
    \label{fig:mnistuspssensitivity} 
    \includegraphics[width=0.48\columnwidth]{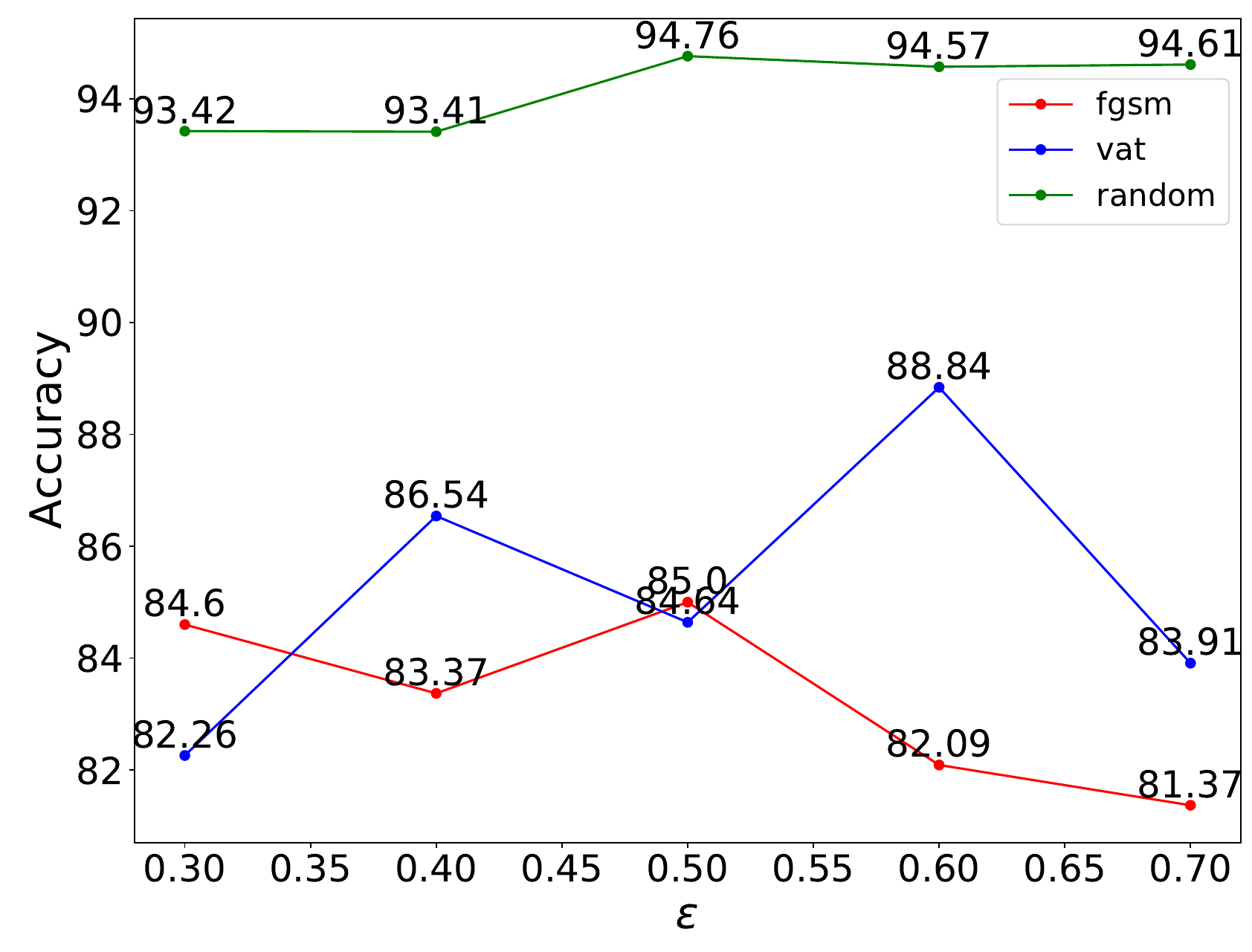}}
  \subfigure[SYNSIG$\rightarrow$GTSRB]{
    \label{fig:synthgtrsbsensitivity} 
    \includegraphics[width=0.48\columnwidth]{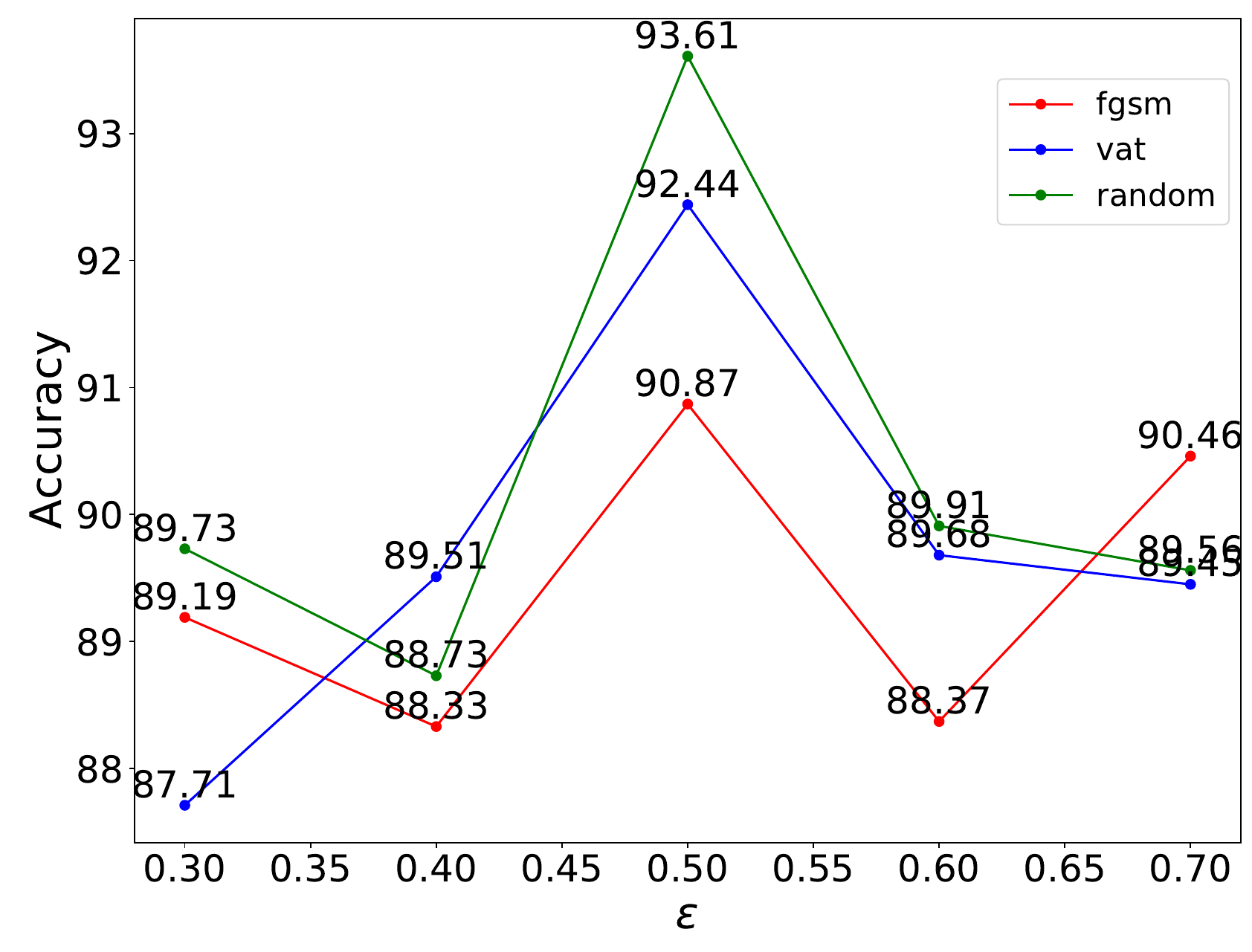}}
	\subfigure[SVHN$\rightarrow$MNIST]{
    \label{fig:svhnmnistsensitivity} 
    \includegraphics[width=0.48\columnwidth]{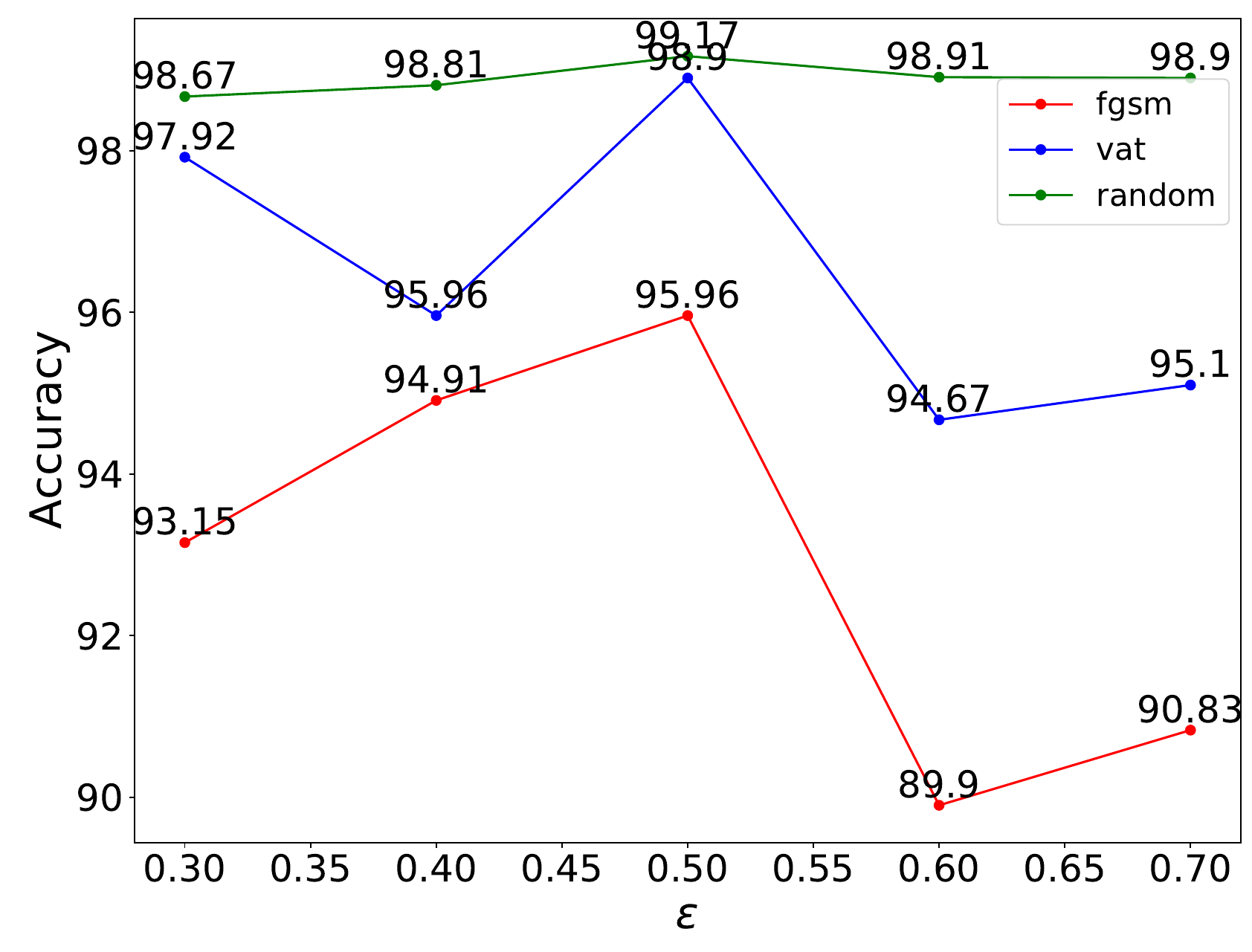}}
\caption{(a)-(d): We test different $\epsilon\in\{0.3, 0.4, 0.5, 0.6, 0.7\}$ on the digits datasets. SRDA$^F$, SRDA$^V$ and SRDA$^G$ are denoted by fgsm, vat and random, respectively}
  \label{fig:sensitivity} 
\end{figure}

As shown in Fig.~\ref{fig:sensitivity}, three models of SRDA are robust in most settings. In particular, SRDA$^G$'s accuracy fluctuates no more than 5\% among the four tasks. Moreover, except SYNSIG$\rightarrow$GTSRB, SRDA$^G$'s accuracy fluctuates no more than 1.5\%. SRDA$^G$ shows robust performance as $\epsilon$ varies, thus meaning that there is no need to tune hyper-parameters subtly for it. SRDA$^V$ performs stably in USPS$\rightarrow$MNIST and SVHN$\rightarrow$MNIST whereas its accuracy varies by more than 4.5\% in MNIST$\rightarrow$USPS and SYNSIG$\rightarrow$GTSRB. SRDA$^F$ holds stability in USPS$\rightarrow$MNIST, MNIST$\rightarrow$USPS and SYNSIG$\rightarrow$GTSRB. In SVHN$\rightarrow$MNIST, its accuracy varies by more than 6\%. The experimental results confirm the belief that SRDA is robust in most settings as $\epsilon$ changes. Especially, SRDA$^G$ is the most stable one. 

\subsubsection{Discussion of Local Smooth Discrepancy}

\begin{figure}[h]
  \centering
  \subfigure[SRDA$^G$]{
    \label{fig:randomlossacc} 
    \includegraphics[width=0.48\columnwidth]{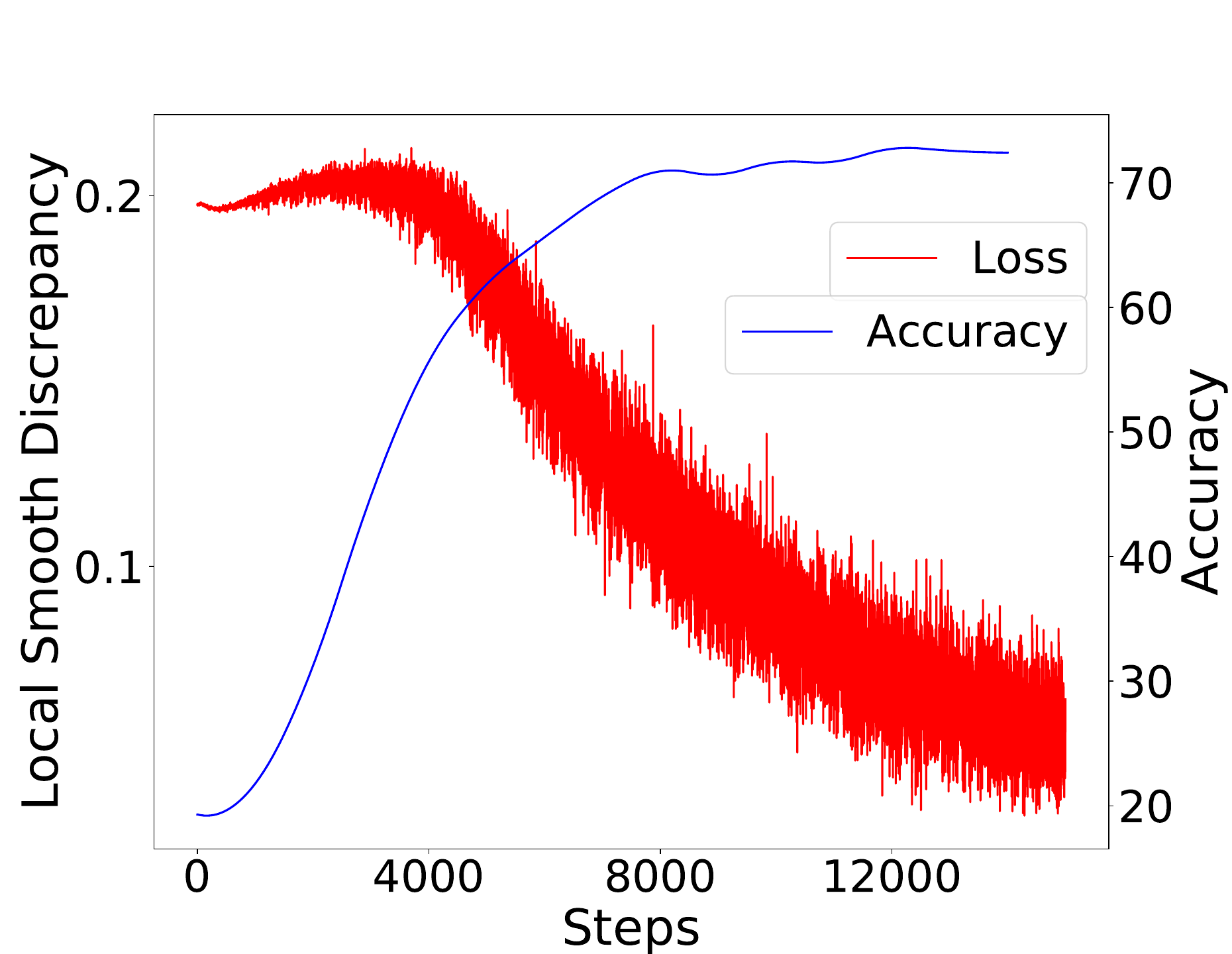}}
  \subfigure[SRDA$^F$]{
    \label{fig:fgsmlossacc} 
    \includegraphics[width=0.48\columnwidth]{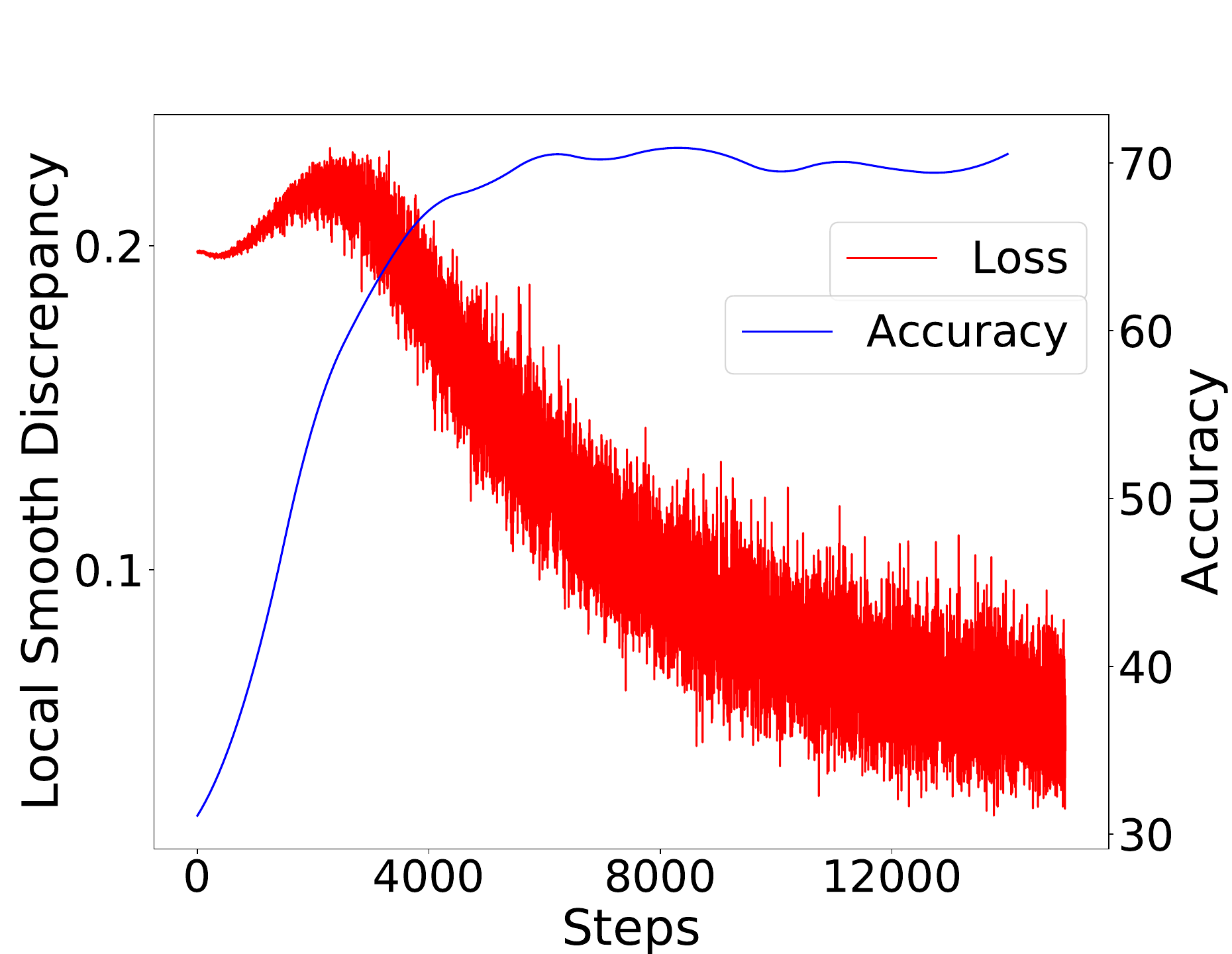}}
  \subfigure[SRDA$^V$]{
    \label{fig:vatlossacc} 
    \includegraphics[width=0.48\columnwidth]{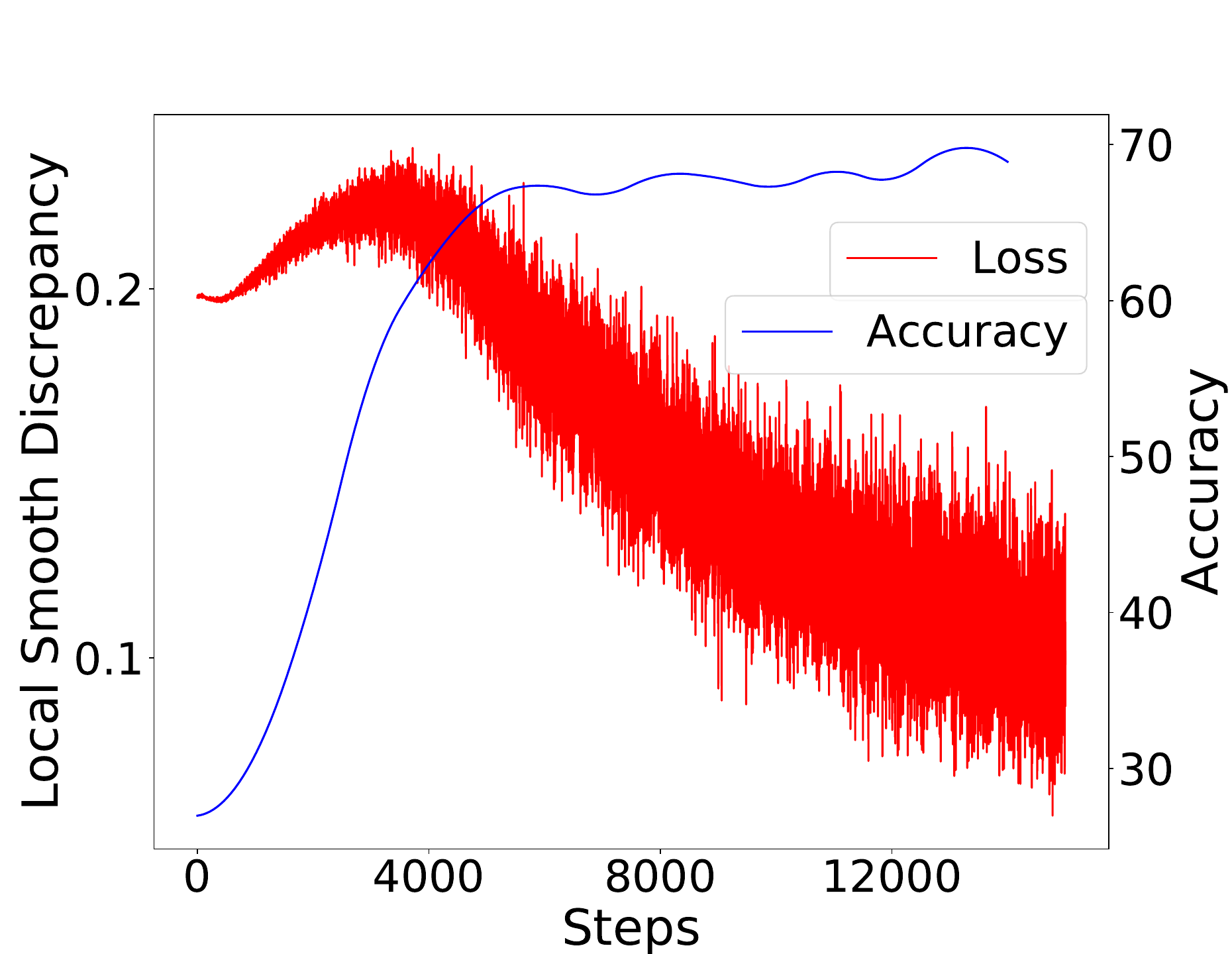}}
	\subfigure[MCD]{
    \label{fig:lossaccMCD} 
    \includegraphics[width=0.48\columnwidth]{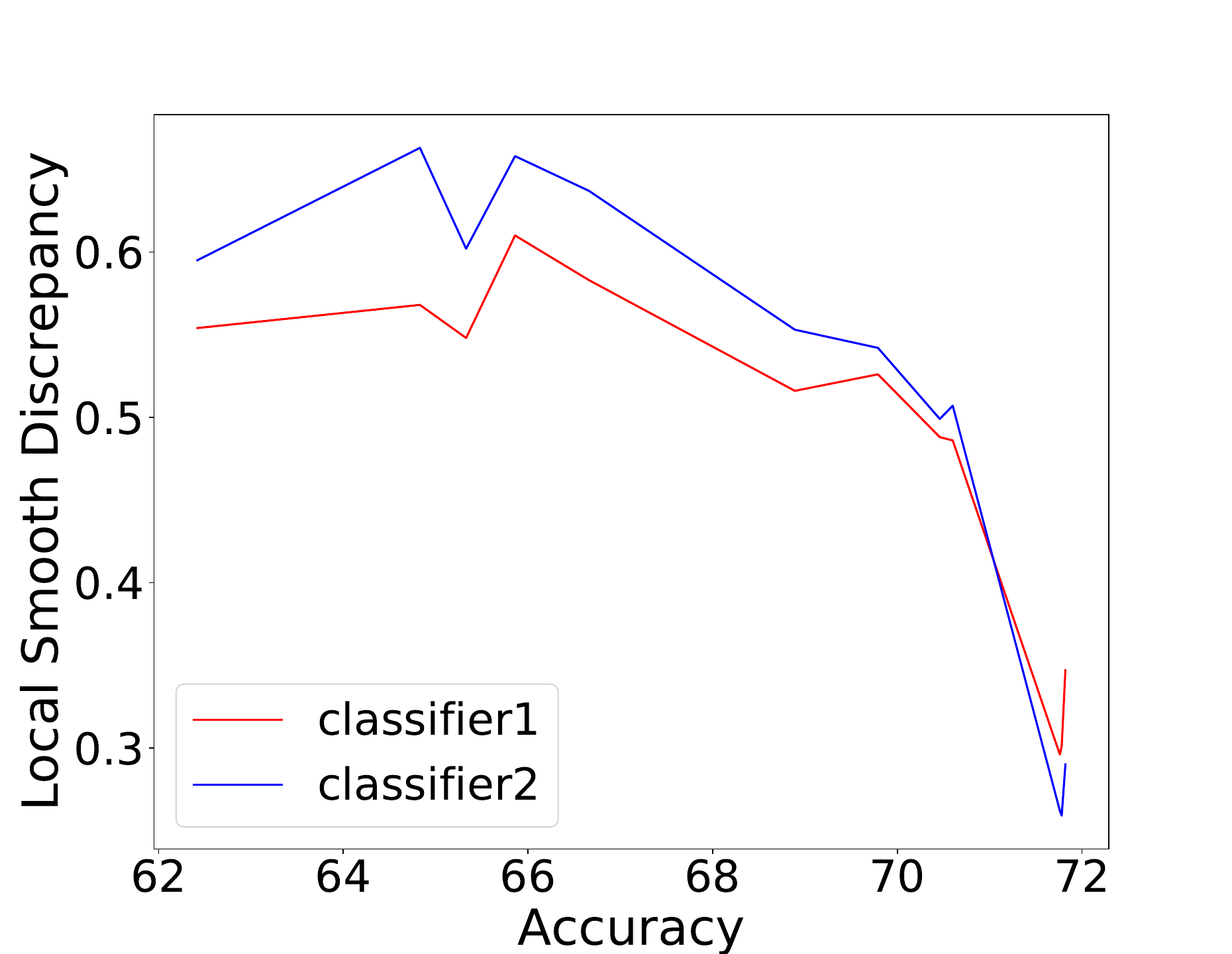}} 
\caption{(a)-(c) display the relationship between LSD (red line) and a model's accuracy (blue line). Three SRDA models are evaluated on VisDA. As discrepancy decreases, the accuracy increases. (d) displays the relationship between LSD and a model's accuracy in MCD. The model with higher accuracy gets a lower LSD.}
  \label{fig:lossacc} 
\end{figure}

To verify that LSD reflects the performance of a model, we show the relationship between LSD and a model's accuracy in Fig.~\ref{fig:lossacc}. Three models, i.e., SRDA$^F$, SRDA$^V$ and SRDA$^G$, are assessed on VisDA. Because we get an accuracy value every epoch and LSD values are recorded every step,
we apply a quadratic interpolation for accuracy values.

As shown in Figs.~\ref{fig:randomlossacc}-(c), the accuracy of all three models gradually increases as LSD decreases. Further, SRDA$^G$ with the highest accuracy refers to the lowest LSD and SRDA$^F$ with the lowest accuracy refers to the highest LSD. The relationship between an accuracy value and LSD indicates that LSD is a reasonable metric to evaluate the performance of a UDA model. This is a remarkable property because adversarial training that conducts a min-max optimization lacks a metric to supervise a training procedure~\cite{ganin2016domain}. Its loss value shows no obvious relationship with its accuracy. 

Moreover, to prove that LSD is a useful metric to assess the performance of a UDA model, we further test it on MCD. We train 12 MCD models with different accuracy on VisDA by tuning hyper-parameters. We generate adversarial samples with SRDA$^{F*}$ to ensure fairness. To ensure valid results, both classifiers in MCD are tested.

As shown in Fig.~\ref{fig:lossaccMCD}, LSD and accuracy are negatively correlated. We train 12 MCD models with accuracy percentages belong to $\{62.42\%, 64.83\%, 65.33\%, 65.86\%,$ $ 66.66\%, 68.89\%, 69.79\%, 70.46\%, 70.60\%, 71.76\%, 71.78\%, $ $71.82\%\}$. LSD of both classifiers decrease from 0.6 to 0.3 roughly as models' accuracy increases. Because FGSM is based on a gradient descent algorithm, randomness of such algorithm causes some fluctuations that several MCD models with high accuracy show relatively high LSD, such as the model with accuracy of 71.82\%.  Overall, LSD can be a proper metric to evaluate the performance of a UDA method.

\subsubsection{Limitation}
This work answers why a local-Lipschitz constraint is beneficial to the solution of a UDA problem and proposes a novel Lipschitz-constraint-based method that considers the effect of the sample amount of a target domain, and the dimension and batchsize of samples. A limitation of the proposed method needs to be noted. As shown in the Office-31 classification experiment, both SRDA$^F$ and SRDA$^G$ get inferior performance. A large amount of the target domain is necessary for Lipschitz-constraint-based methods. Because Office-31 consists of 4,110 images only, the proposed method cannot detect enough sensitive samples to form a large margin between  a decision boundary and samples. Thus, it shows poor performance on Office-31.  The results demonstrate that the proposed method fails to handle well a small amount of data. To make it work well with such limited data cases, exploring a better method than the proposed one to detect more sensitive samples is our future work.

\section{Conclusion}
In this paper, we have proposed a method for UDA inspired by a probabilistic Lipschitz constraint. Principles analyzed in~\cite{Ben-David2014} are well extended to a deep end-to-end model and a practical optimization strategy is presented. The key to strengthening Lipschitz continuity is to minimize a local smooth discrepancy defined in this paper. To avoid a loose error bound, the proposed optimization strategy is subtly designed by considering the dimension and batchsize of samples, which have been ignored by previous Lipschitz-constraint-based methods
~\cite{shu2018a,miyato2018virtual,mao2019virtual}. 
Experiments demonstrate that a  Lipschitz constraint without adversarial training is effective for UDA and factors we discuss are critical to an efficient and stable UDA model. Our future work includes seeking  theoretically tighter error bounds and applications of the proposed methods to security, manufacturing and transportation 
\cite{9049451,9205684,8951126,8848867}.


%





\ifCLASSOPTIONcaptionsoff
  \newpage
\fi

\bibliographystyle{IEEEtran}
\bibliography{srda}

\end{document}